\documentclass[10pt,journal,compsoc,twoside]{IEEEtran}

\usepackage{graphicx} \graphicspath{{img/}}
\usepackage[nocompress]{cite}
\usepackage{amsmath,amsfonts}
\usepackage{microtype}
\usepackage{booktabs}
\usepackage{balance}

\usepackage[pdftex, bookmarksnumbered=true, pagebackref=false]{hyperref}
\hypersetup{colorlinks,
citecolor=[rgb]{0.0,0.0,0.4},
filecolor=[rgb]{0.6,0.0,0.4},
linkcolor=[rgb]{0.0,0.0,0.4},
urlcolor=[rgb]{0.0,0.0,0.4}}

\def\FP#1{{\color{red} [FP: {\it{#1}} ]}}
\def\SM#1{{\color{blue} [SM: {\it{#1}} ]}}

\begin{document}

\title{Image Segmentation Using Deep Learning:\\ A Survey}

\author{Shervin~Minaee, 
        Yuri~Boykov, 
        Fatih~Porikli, 
        Antonio~Plaza, 
        Nasser~Kehtarnavaz, 
        and Demetri~Terzopoulos
        }

\IEEEtitleabstractindextext{%
\begin{abstract}

Image segmentation is a key topic in image processing and computer vision with applications such as scene understanding, medical image analysis, robotic perception, video surveillance, augmented reality, and image compression, among many others. Various algorithms for image segmentation have been developed in the literature. Recently, due to the success of deep learning models in a wide range of vision applications, there has been a substantial amount of works aimed at developing image segmentation approaches using deep learning models. In this survey, we provide a comprehensive review of the literature at the time of this writing, covering a broad spectrum of pioneering works for semantic and instance-level segmentation, including fully convolutional pixel-labeling networks, encoder-decoder architectures, multi-scale and pyramid based approaches, recurrent networks, visual attention models, and generative models in adversarial settings. We investigate the similarity, strengths and challenges of these deep learning models, examine the most widely used datasets, report performances, and discuss promising future research directions in this area. 
\end{abstract}

\begin{IEEEkeywords}
Image segmentation, deep learning, convolutional neural networks, encoder-decoder models, recurrent models, generative models, semantic segmentation, instance segmentation, medical image segmentation.
\end{IEEEkeywords}}

\maketitle

\IEEEdisplaynontitleabstractindextext

\IEEEraisesectionheading{\section{Introduction}
\label{sec:introduction}}

\begin{figure}
\centering
\includegraphics[page=1,width=0.8\linewidth]{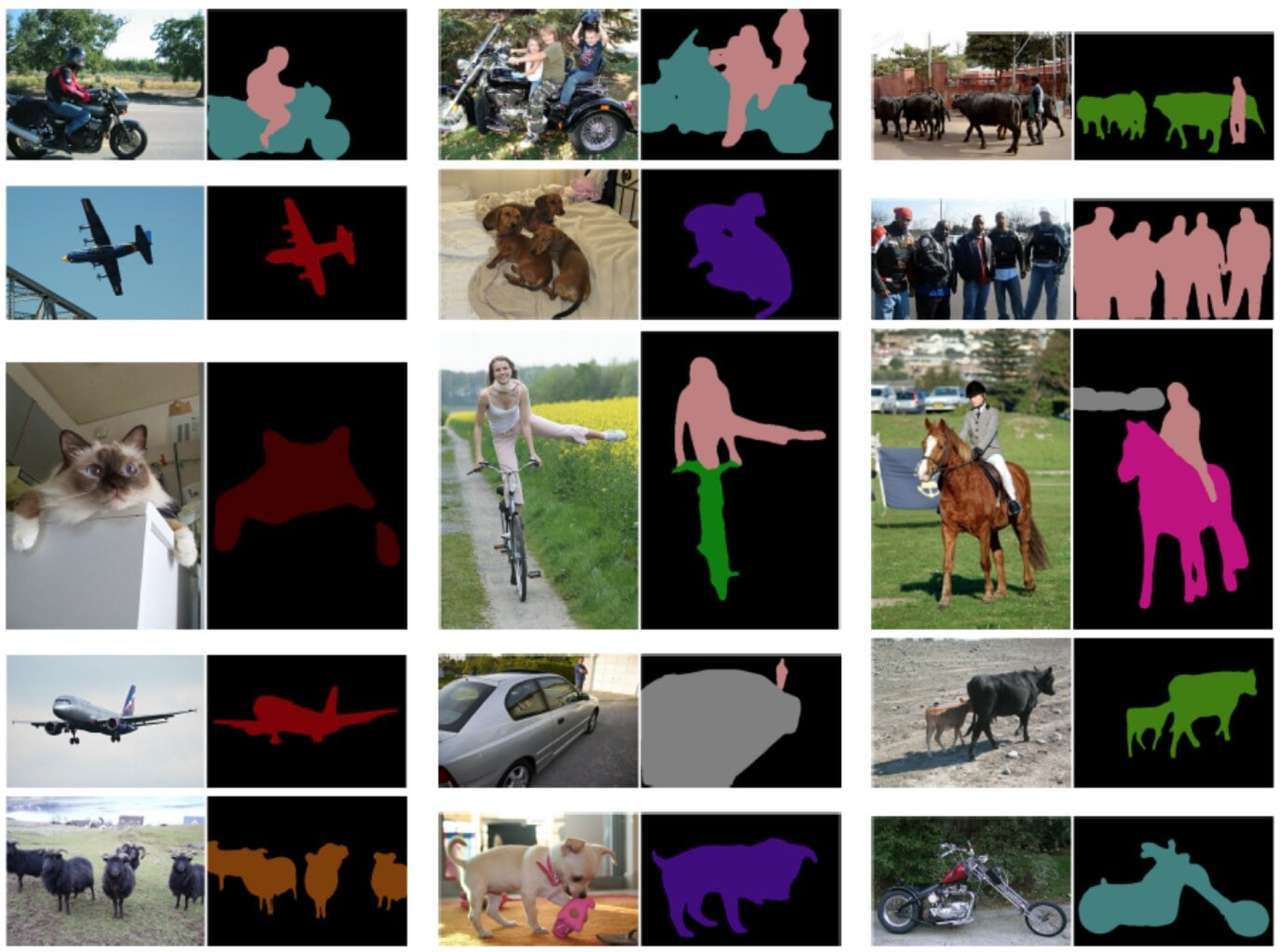}
\caption{Segmentation results of DeepLabV3 \cite{deeplabv3} on sample images.}
\label{intro_sample_results}
\end{figure}

\IEEEPARstart{I}{mage} segmentation is an essential component in many visual understanding systems. It involves partitioning images (or video frames) into multiple segments or objects \cite{vision_book1}. Segmentation plays a central role in a broad range of applications \cite{vision_book2}, including medical image analysis (e.g., tumor boundary extraction and measurement of tissue volumes), autonomous vehicles (e.g., navigable surface and pedestrian detection), video surveillance, and augmented reality to count a few. Numerous image segmentation algorithms have been developed in the literature, from the earliest methods, such as thresholding \cite{otsu1979threshold}, histogram-based bundling, region-growing \cite{region_grow}, k-means clustering \cite{seg_clus}, watersheds \cite{najman1994watershed}, to more advanced algorithms such as active contours \cite{Snakes}, graph cuts \cite{graphcut}, conditional and Markov random fields \cite{plath2009multi}, and sparsity-based \cite{seg_sparse}-\cite{seg_sparse2} methods. Over the past few years, however, deep learning (DL) models have yielded a new generation of image segmentation models with remarkable performance improvements ---often achieving the highest accuracy rates on popular benchmarks--- resulting in 
a paradigm shift in the field. For example, Figure~\ref{intro_sample_results} presents image segmentation outputs of a popular deep learning model, DeepLabv3 \cite{deeplabv3}.

Image segmentation can be formulated as a classification problem of pixels with semantic labels (semantic segmentation) or partitioning of individual objects (instance segmentation). Semantic segmentation performs pixel-level labeling with a set of object categories (e.g., human, car, tree, sky) for all image pixels, thus it is generally a harder undertaking than image classification, which predicts a single label for the entire image. Instance segmentation extends semantic segmentation scope further by detecting and delineating each object of interest in the image (e.g., partitioning of individual persons).

Our survey covers the most recent literature in image segmentation and discusses more than a hundred deep learning-based segmentation methods proposed until 2019. 
We provide a comprehensive review and insights on different aspects of these methods, including the training data, the choice of network architectures, loss functions, training strategies, and their key contributions. We present a comparative summary of the performance of the reviewed methods and discuss several challenges and potential future directions for deep learning-based image segmentation models.

We group deep learning-based works into the following categories based on their main technical contributions: 

\begin{enumerate}
    \item Fully convolutional networks
    \item Convolutional models with graphical models
    \item Encoder-decoder based models
    \item Multi-scale and pyramid network based models
    \item R-CNN based models (for instance segmentation)
    \item Dilated convolutional models and DeepLab family
    \item Recurrent neural network based models
    \item Attention-based models
    \item Generative models and adversarial training
    \item Convolutional models with active contour models
    \item Other models
\end{enumerate}

Some the key contributions of this survey paper can be summarized as follows:
\begin{itemize}
 \item This survey covers the contemporary literature with respect
 to segmentation problem, and overviews more than 100 segmentation algorithms proposed till 2019, grouped into 10 categories.
 \item We provide a comprehensive review and an insightful analysis of different aspects of segmentation algorithms using deep learning, including the training data, the choice of network architectures, loss functions, training strategies, and their key contributions.
 \item We provide an overview of around 20 popular image segmentation datasets, grouped into 2D, 2.5D (RGB-D), and 3D images.
 \item We provide a comparative summary of the properties and performance of the reviewed methods for segmentation purposes, on popular benchmarks. 
 \item We provide several challenges and potential future directions for deep learning-based image segmentation.
\end{itemize}

The remainder of this survey is organized as follows:
Section~\ref{sec:DNNs} provides an overview of popular deep neural
network architectures that serve as the backbone of many
modern segmentation algorithms. 
Section~\ref{sec:dl-models} provides a comprehensive overview of the
most significant state-of-the-art deep learning based segmentation
models, more than 100 till 2020. 
We also discuss their strengths and contributions over previous works here.
Section~\ref{sec:datasets}
reviews some of the most popular image segmentation datasets and their
characteristics. 
Section~\ref{sec:metrics} reviews popular metrics
for evaluating deep-learning-based segmentation models.
In Section~\ref{sec:quant_result}, we report the quantitative results
and experimental performance of these models. In Section~\ref{sec:challenges}, we discuss the main challenges and
future directions for deep learning-based segmentation methods.
Finally, we present our conclusions in Section~\ref{sec:conclusions}.


\section{Overview of Deep Neural Networks}
\label{sec:DNNs}

This section provides an overview of some of the most prominent deep
learning architectures used by the computer vision community,
including convolutional neural networks (CNNs) \cite{CNN}, recurrent
neural networks (RNNs) and long short term memory (LSTM) \cite{lstm},
encoder-decoders \cite{segnet}, and generative adversarial networks
(GANs) \cite{GAN}. With the popularity of deep learning in recent
years, several other deep neural architectures have been proposed,
such as transformers, capsule networks, gated recurrent units,
spatial transformer networks, etc., which will not be covered here.

It is worth mentioning that in some cases the DL-models can be trained from scratch on new applications/datasets (assuming a sufficient quantity of labeled training data), but in many cases there are not enough labeled data available to train a model from scratch and one can use \textbf{transfer learning}  to tackle this problem.
In transfer learning, a model trained on one task is re-purposed on another (related) task, usually by some adaptation process toward the new task. 
For example, one can imagine adapting an image classification model trained on ImageNet to a different task, such as texture classification, or face recognition.
In image segmentation case, many people use a model trained on ImageNet (a larger dataset than most of image segmentation datasets), as the encoder part of the network, and re-train their model from those initial weights. 
The assumption here is that those pre-trained models should be able to capture the semantic information of the image required for segmentation, and therefore enabling them to train the model with less labeled samples.

\subsection{Convolutional Neural Networks (CNNs)}

\begin{figure}[h]
\centering
\includegraphics[page=3,width=0.98\linewidth]{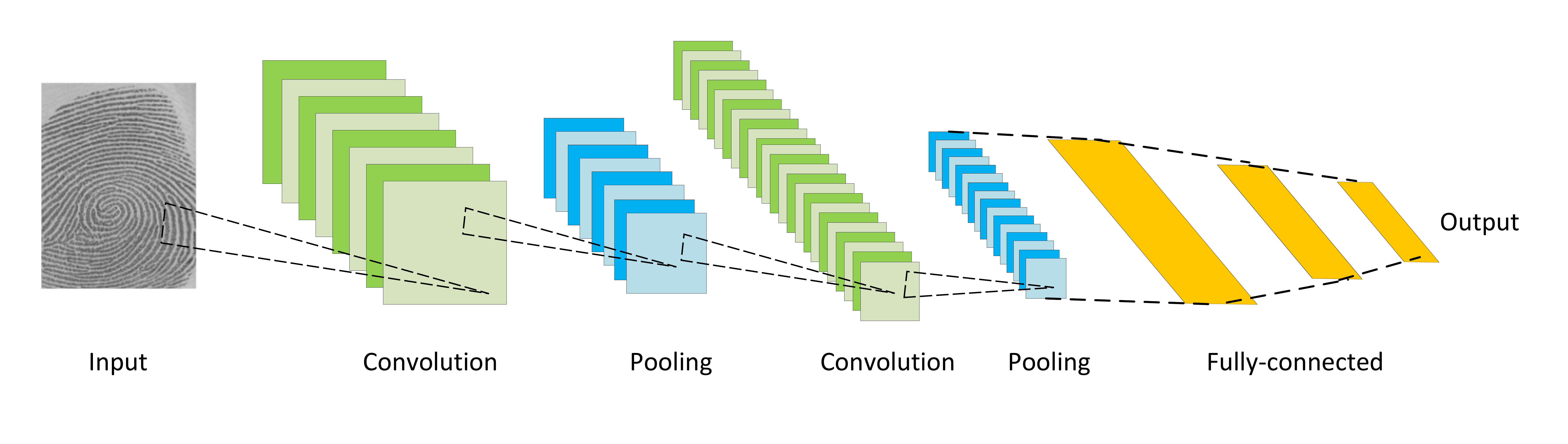}
\caption{Architecture of convolutional neural networks. From
\cite{CNN}.}
\label{fig:CNN_arch}
\end{figure}

CNNs are among the most successful and widely used architectures in
the deep learning community, especially for computer vision tasks.
CNNs were initially proposed by Fukushima in his seminal
paper on the ``Neocognitron'' \cite{neocog}, based on the hierarchical receptive
field model of the visual cortex proposed by Hubel and Wiesel.
Subsequently, Waibel \textit{et al.} \cite{Waibel} introduced CNNs with weights shared among temporal receptive fields and backpropagation
training for phoneme recognition, and LeCun \textit{et al.} \cite{CNN}
developed a CNN architecture for document recognition
(Figure~\ref{fig:CNN_arch}).

CNNs mainly consist of three type of layers: i) convolutional layers,
where a kernel (or filter) of weights is convolved in order to extract
features; ii) nonlinear layers, which apply an activation function on
feature maps (usually element-wise) in order to enable the modeling of
non-linear functions by the network; and iii) pooling layers, which
replace a small neighborhood of a feature map with some statistical
information (mean, max, etc.) about the neighborhood and reduce
spatial resolution. The units in layers are locally connected; that
is, each unit receives weighted inputs from a small neighborhood,
known as the receptive field, of units in the previous layer. By
stacking layers to form multi-resolution pyramids, the higher-level
layers learn features from increasingly wider receptive fields. The
main computational advantage of CNNs is that all the receptive fields
in a layer share weights, resulting in a significantly smaller number
of parameters than fully-connected neural networks.
Some of the most well-known CNN architectures include: AlexNet \cite{alexnet}, VGGNet
\cite{vggnet}, ResNet \cite{resnet}, GoogLeNet \cite{googlenet},
MobileNet \cite{mobilenet}, and DenseNet \cite{densenet}.

\subsection{Recurrent Neural Networks (RNNs) and the LSTM}

\begin{figure}[h]
\centering
\includegraphics[page=1,width=0.8\linewidth]{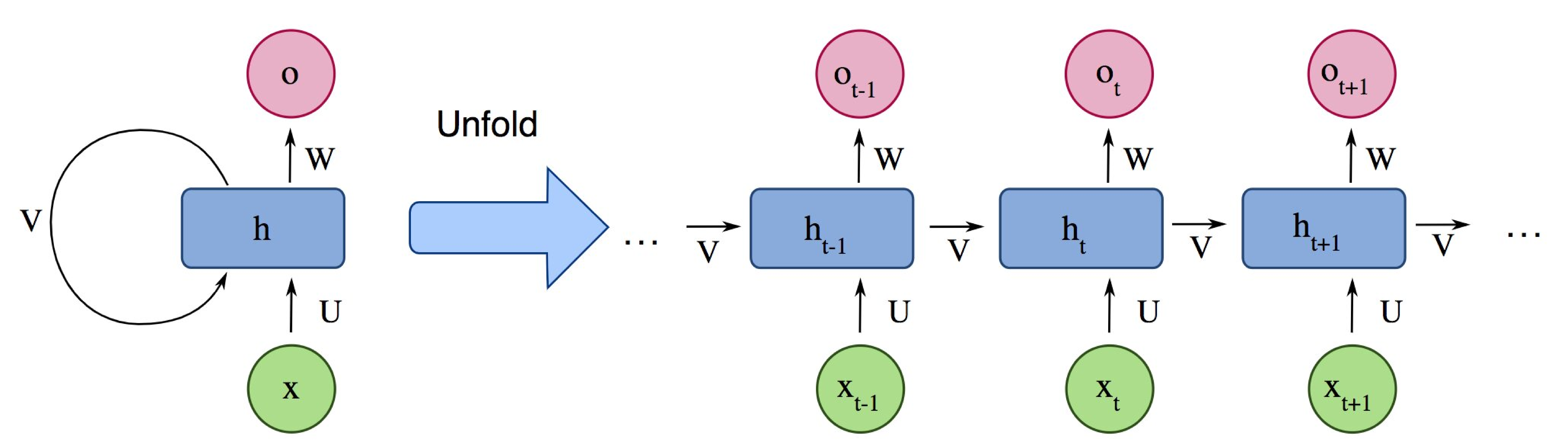}
\caption{Architecture of a simple recurrent neural network.}
\label{fig:RNN_arch}
\end{figure}

RNNs \cite{RNN} are widely used to process sequential data, such as
speech, text, videos, and time-series, where data
at any given time/position depends on previously encountered data. At
each time-stamp the model collects the input from the current time
$X_i$ and the hidden state from the previous step $h_{i-1}$, and
outputs a target value and a new hidden state
(Figure~\ref{fig:RNN_arch}). 

\begin{figure}[h]
\centering
\includegraphics[page=1,width=0.99\linewidth]{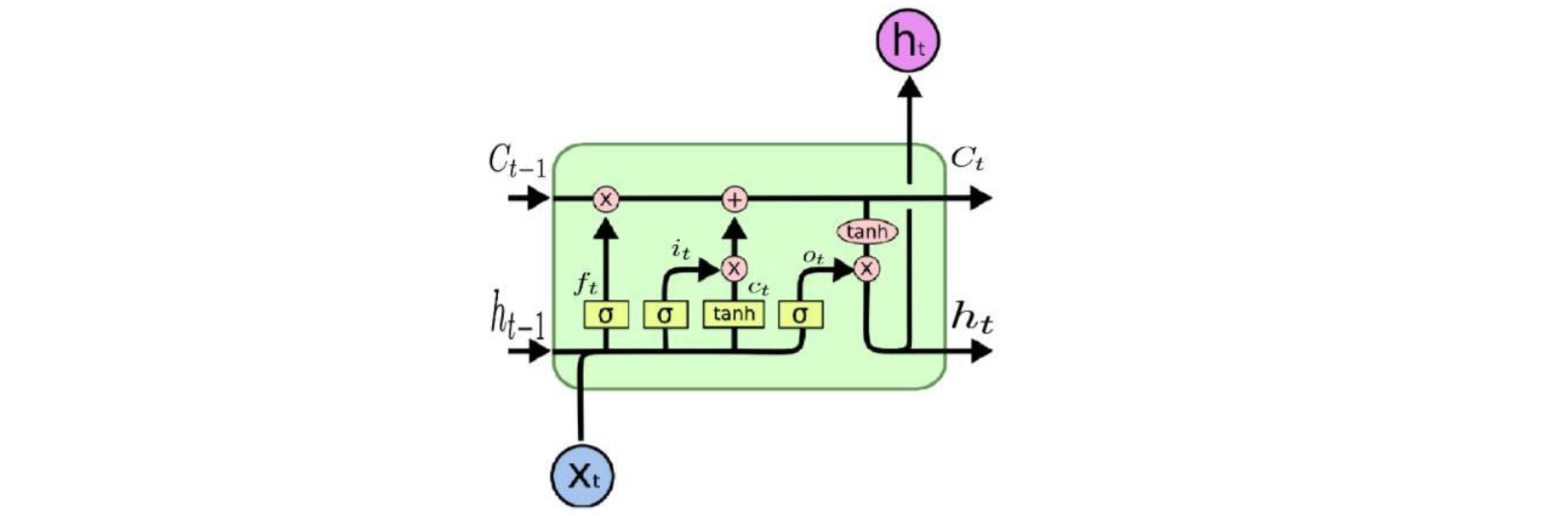}
\caption{Architecture of a standard LSTM module. Courtesy of
Karpathy.}
\label{fig:lstm_model}
\end{figure}

RNNs are typically problematic with long sequences as they cannot
capture long-term dependencies in many real-world applications
(although they exhibit no theoretical limitations in this regard) and
often suffer from gradient vanishing or exploding problems. However, a
type of RNNs called Long Short Term Memory (LSTM) \cite{lstm} is
designed to avoid these issues.
The LSTM architecture (Figure~\ref{fig:lstm_model}) includes three
gates (input gate, output gate, forget gate), which regulate the flow
of information into and out from a memory cell, which stores values
over arbitrary time intervals.

\subsection{Encoder-Decoder and Auto-Encoder Models}

\begin{figure}[h]
\centering
\includegraphics[page=1,width=0.99\linewidth]{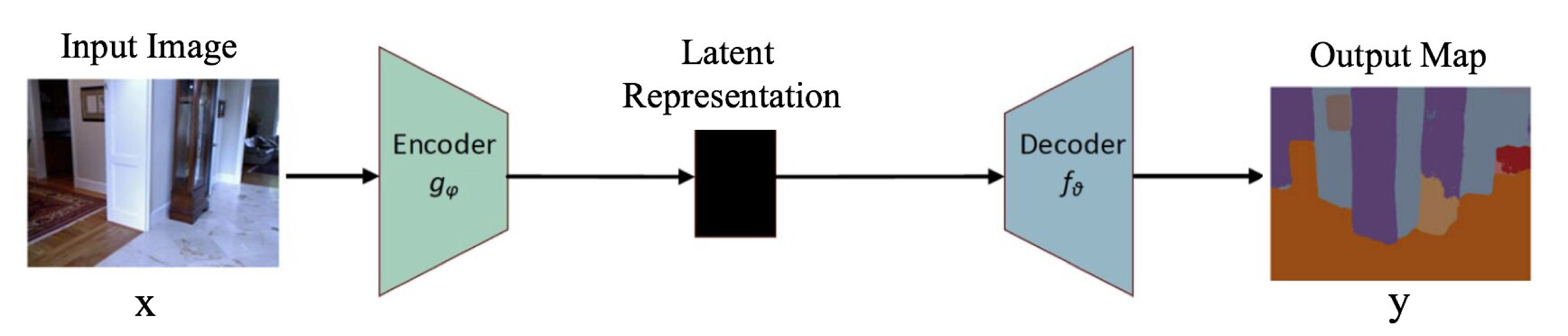}
\caption{The architecture of a simple encoder-decoder model.}
\label{fig:autoencoder}
\end{figure}

Encoder-Decoder models are a family of models which learn to map data-points from an input domain to an output domain via a two-stage network: The encoder, represented
by an encoding function $z=f(x)$, compresses the input into a latent-space
representation; the decoder, $y=g(z)$, aims to predict the output from the latent space representation \cite{DL_book, segnet}.
The latent representation here essentially refers to a feature (vector) representation, which is able to capture the underlying semantic information of the input that is useful for predicting the output.
These models are extremely popular in image-to-image translation problems, as well as for sequence-to-sequence models in NLP.
Figure~\ref{fig:autoencoder} illustrates the block-diagram of a simple encoder-decoder model. 
These models are usually
trained by minimizing the reconstruction loss $L(y,\hat{y})$, which
measures the differences between the ground-truth output $y$ and the
subsequent reconstruction $\hat{y}$.
The output here could be an enhanced version of the image (such as in image de-blurring, or super-resolution), or a segmentation map.
Auto-encoders are special case of encoder-decoder models in which the input and output are the same.

\subsection{Generative Adversarial Networks (GANs)}

\begin{figure}[h]
\centering
\includegraphics[width=0.7\linewidth]{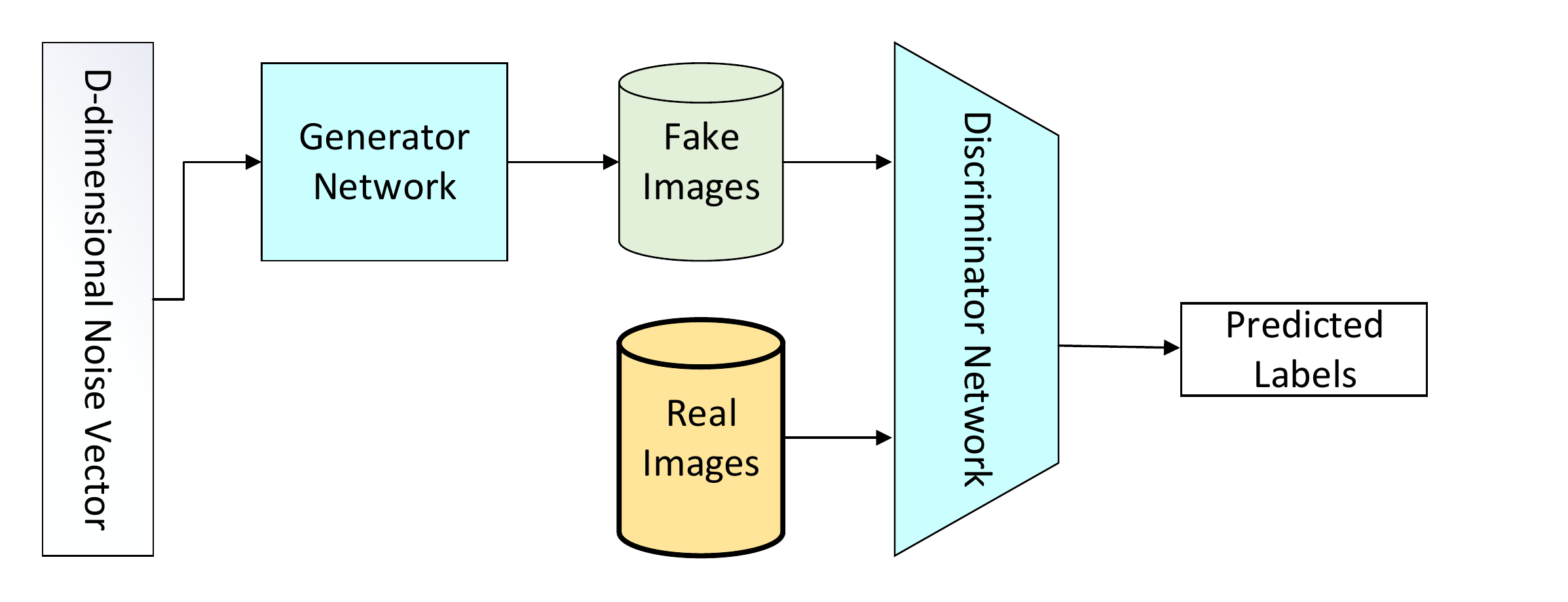}
\caption{Architecture of a generative adversarial network.}
\label{fig:gen_arch}
\end{figure}

GANs are a newer family of deep learning models \cite{GAN}. They
consist of two networks---a generator and a discriminator
(Figure~\ref{fig:gen_arch}). The generator network $G= z \rightarrow
y$ in the conventional GAN learns a mapping from noise $z$ (with a
prior distribution) to a target distribution $y$, which is similar to
the ``real'' samples. The discriminator network $D$ attempts to
distinguish the generated samples (``fakes'') from the ``real'' ones.
The GAN loss function may be written as $\mathcal{L}_\text{GAN}=
\mathbb{E}_{x \sim p_\text{data}(x)}[\log D(x)]+ \mathbb{E}_{z \sim
p_z(z)}[\log(1-D(G(z)))]$. We can regard the GAN as a minimax game
between $G$ and $D$, where $D$ is trying to minimize its
classification error in distinguishing fake samples from real ones,
hence maximizing the loss function, and $G$ is trying to maximize the
discriminator network's error, hence minimizing the loss function.
After training the model, the trained generator model would be $G^*=
\text{arg} \ \min_G \max_D \ \mathcal{L}_\text{GAN}$ In practice, this
function may not provide enough gradient for effectively training $G$,
specially initially (when $D$ can easily discriminate fake samples
from real ones). Instead of minimizing $\mathbb{E}_{z \sim
p_z(z)}[\log(1-D(G(z)))]$, a possible solution is to train it to
maximize $\mathbb{E}_{z \sim p_z(z)}[\log(D(G(z)))]$.

Since the invention of GANs, researchers have endeavored to
improve/modify GANs several ways. For example, Radford \textit{et al.}
\cite{dc-gan} proposed a convolutional GAN model, which works better
than fully-connected networks when used for image generation. Mirza
\cite{con-gan} proposed a conditional GAN model that can generate
images conditioned on class labels, 
which enables one to generate samples with specified
labels. Arjovsky \textit{et al.} \cite{was-gan} proposed a new loss
function based on the Wasserstein (a.k.a. earth mover's distance) to
better estimate the distance for cases in which the distribution of
real and generated samples are non-overlapping (hence the
Kullback–Leiber divergence is not a good measure of the distance).
For additional works, we refer the reader to \cite{GanZoo}.

\section{DL-Based Image Segmentation Models}
\label{sec:dl-models}

This section provides a detailed review of more than a hundred deep learning-based segmentation methods proposed until 2019, grouped into 10 categories (based on their model architecture).
It is worth mentioning that there are some pieces that are common among many of these works, such as having encoder and decoder parts, skip-connections, multi-scale analysis, and more recently the use of dilated convolution.
Because of this, it is difficult to mention the unique contributions of each work, but easier to group them based on their underlying architectural contribution over previous works.
Besides the architectural categorization of these models, one can also group them based on the segmentation goal into: semantic, instance, panoptic, and depth segmentation categories. But due to the big difference in terms of volume of work in those tasks, we decided to follow the architectural grouping.

\subsection{Fully Convolutional Networks}

\begin{figure}[h]
\centering
\includegraphics[width=0.5\linewidth]{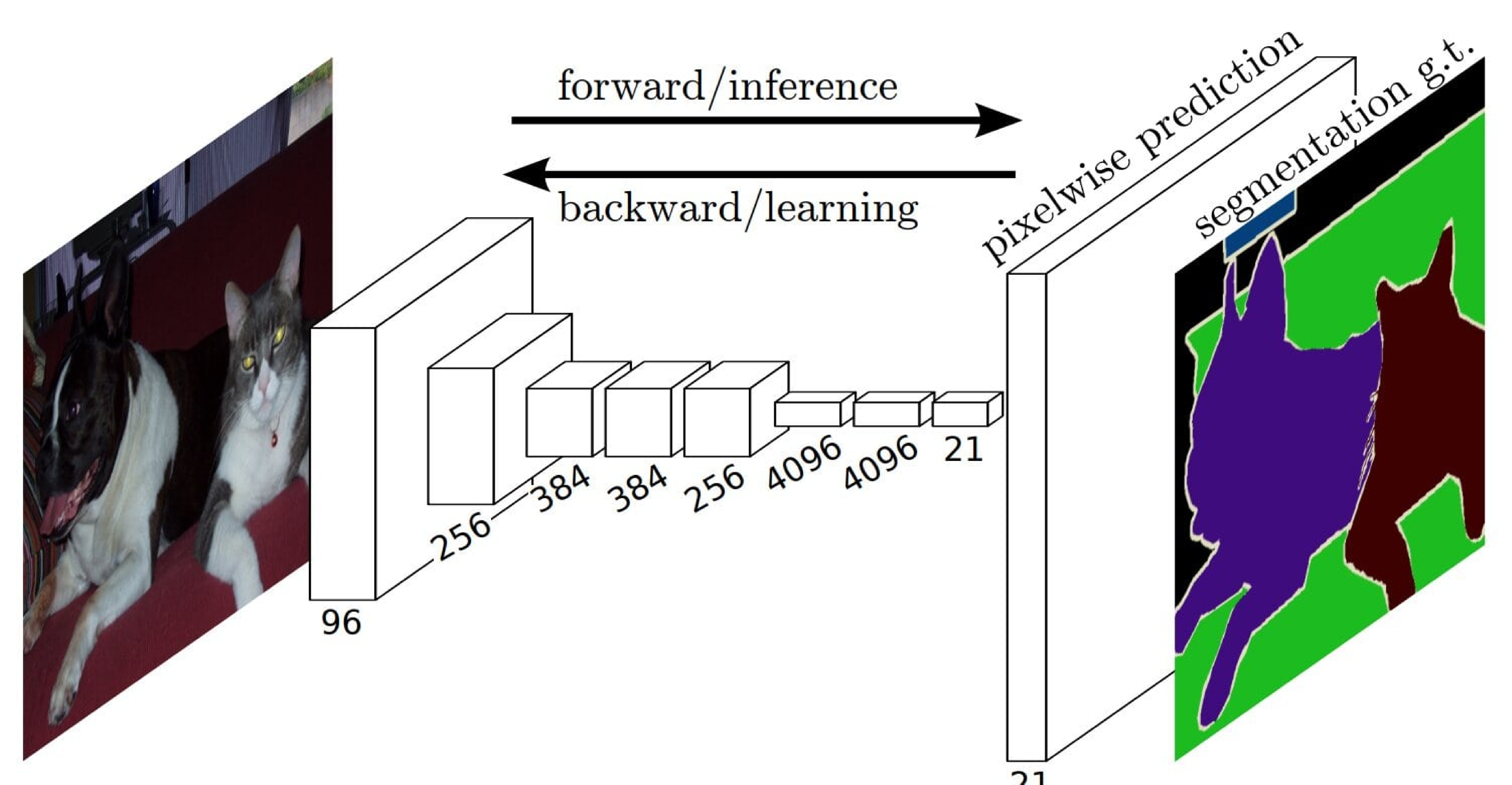}
\caption{A fully convolutional image segmentation network. The FCN
learns to make dense, pixel-wise predictions. From \cite{seg_fcn}.}
\label{fig:FCN_blk}
\end{figure}

Long \textit{et al.} \cite{seg_fcn} proposed one of the first deep
learning works for semantic image segmentation, using a fully
convolutional network (FCN). An FCN (Figure~\ref{fig:FCN_blk})
includes only convolutional layers, which enables it to take an image
of arbitrary size and produce a segmentation map of the same size. The
authors modified existing CNN architectures, such as VGG16 and
GoogLeNet, to manage non-fixed sized input and output, by replacing
all fully-connected layers with the fully-convolutional layers. As a
result, the model outputs a spatial segmentation map instead of
classification scores.

\begin{figure}[h]
\centering
\includegraphics[width=0.8\linewidth]{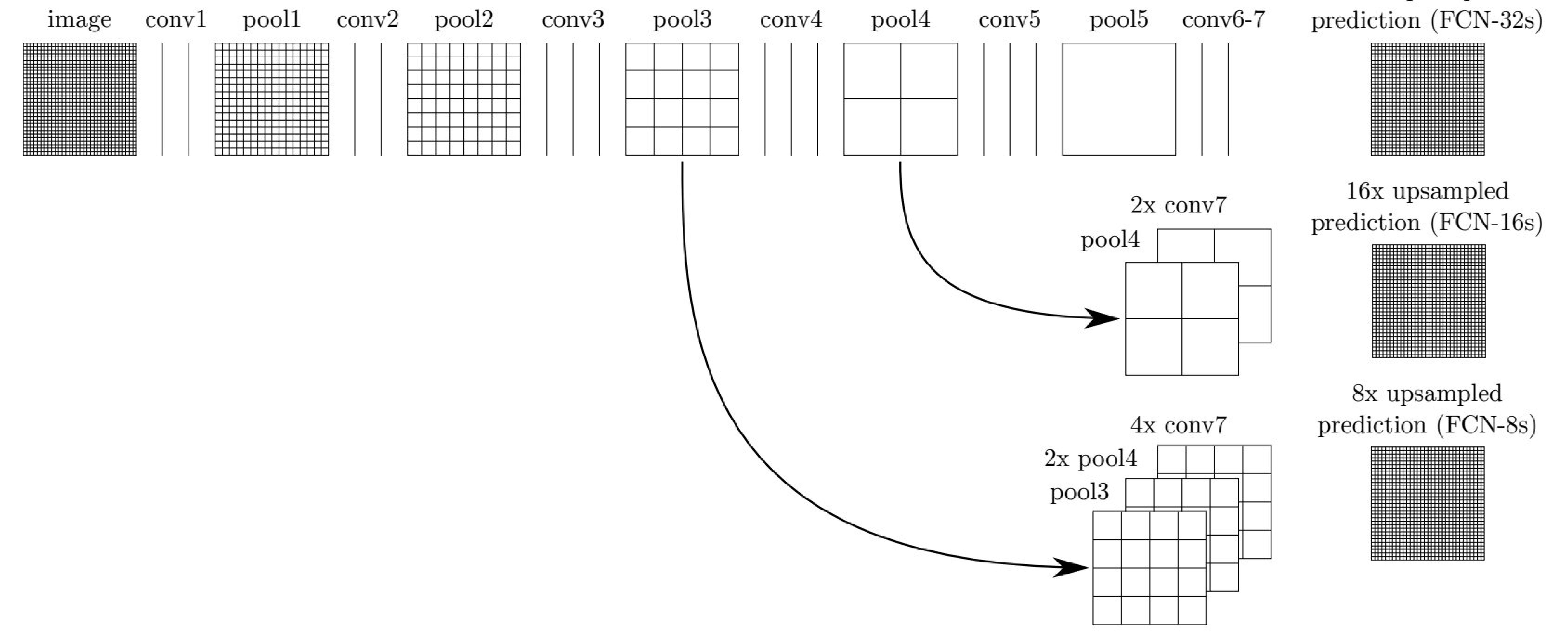}
\caption{Skip connections combine coarse, high-level information and
fine, low-level information. From \cite{seg_fcn}.}
\label{fig:FCN_blk2}
\end{figure}

Through the use of skip connections in which feature maps from the final layers of the model are up-sampled and
fused with feature maps of earlier layers (Figure~\ref{fig:FCN_blk2}),
the model combines semantic information (from deep, coarse layers) and
appearance information (from shallow, fine layers) in order to produce
accurate and detailed segmentations. The model was tested on PASCAL
VOC,
NYUDv2, and SIFT Flow, and achieved state-of-the-art segmentation
performance.

This work is considered a milestone in image segmentation,
demonstrating that deep networks can be trained for semantic
segmentation in an end-to-end manner on variable-sized images.
However, despite its popularity and effectiveness, the conventional
FCN model has some limitations---it is not fast enough for real-time
inference, it does not take into account the global context
information in an efficient way, and it is not easily transferable to 3D images. Several efforts have attempted to overcome some of the
limitations of the FCN.

\begin{figure}[h]
\centering
\includegraphics[width=0.8\linewidth]{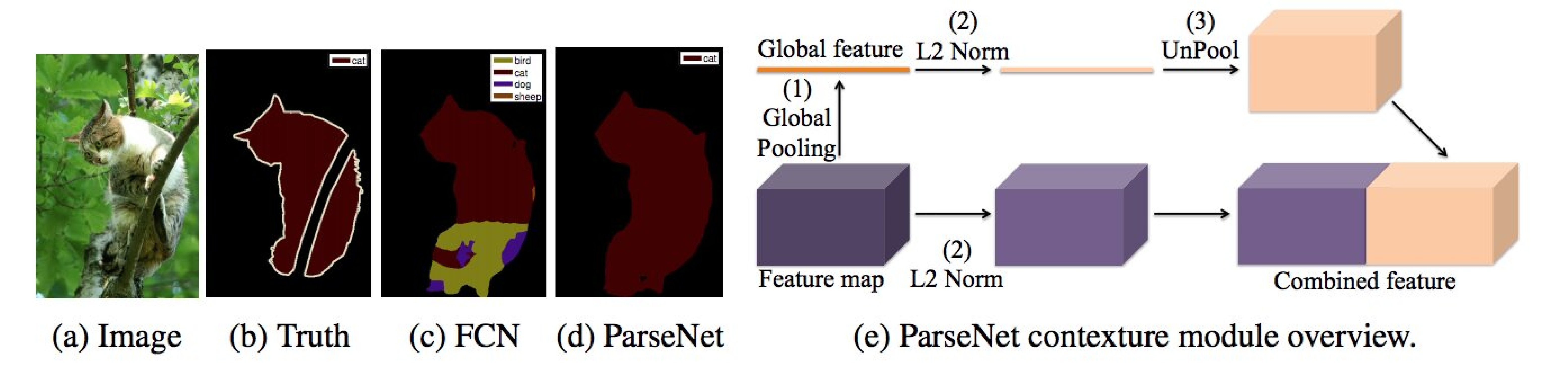}
\caption{ParseNet, showing the use of extra global context to produce
smoother segmentation (d) than an FCN (c). From \cite{seg_parsenet}.}
\label{fig:parse_blk}
\end{figure}

For instance, Liu \textit{et al.} \cite{seg_parsenet} proposed a model
called ParseNet, to address an issue with FCN---ignoring global
context information. ParseNet adds global context to FCNs by using the
average feature for a layer to augment the features at each location.
The feature map for a layer is pooled over the whole image resulting
in a context vector. 
This context vector is normalized
and unpooled to produce new feature maps of the same size as the
initial ones. These feature maps are then concatenated. In a nutshell, ParseNet is an FCN with the described module replacing the convolutional layers (Figure~\ref{fig:parse_blk}).

FCNs have been applied to a variety of segmentation problems, such as brain tumor segmentation \cite{fcn_brain},
instance-aware semantic segmentation \cite{fcn_instance}, skin lesion
segmentation \cite{fcn_skin}, and iris segmentation \cite{fcn_iris}.

\subsection{Convolutional Models With Graphical Models}
\label{sec:cnn+crf}

As discussed, FCN ignores potentially useful scene-level semantic context. To integrate more context, several approaches incorporate probabilistic
graphical models, such as Conditional Random Fields (CRFs) and Markov
Random Field (MRFs), into DL architectures.

\begin{figure}[h]
\centering
\includegraphics[width=0.8\linewidth]{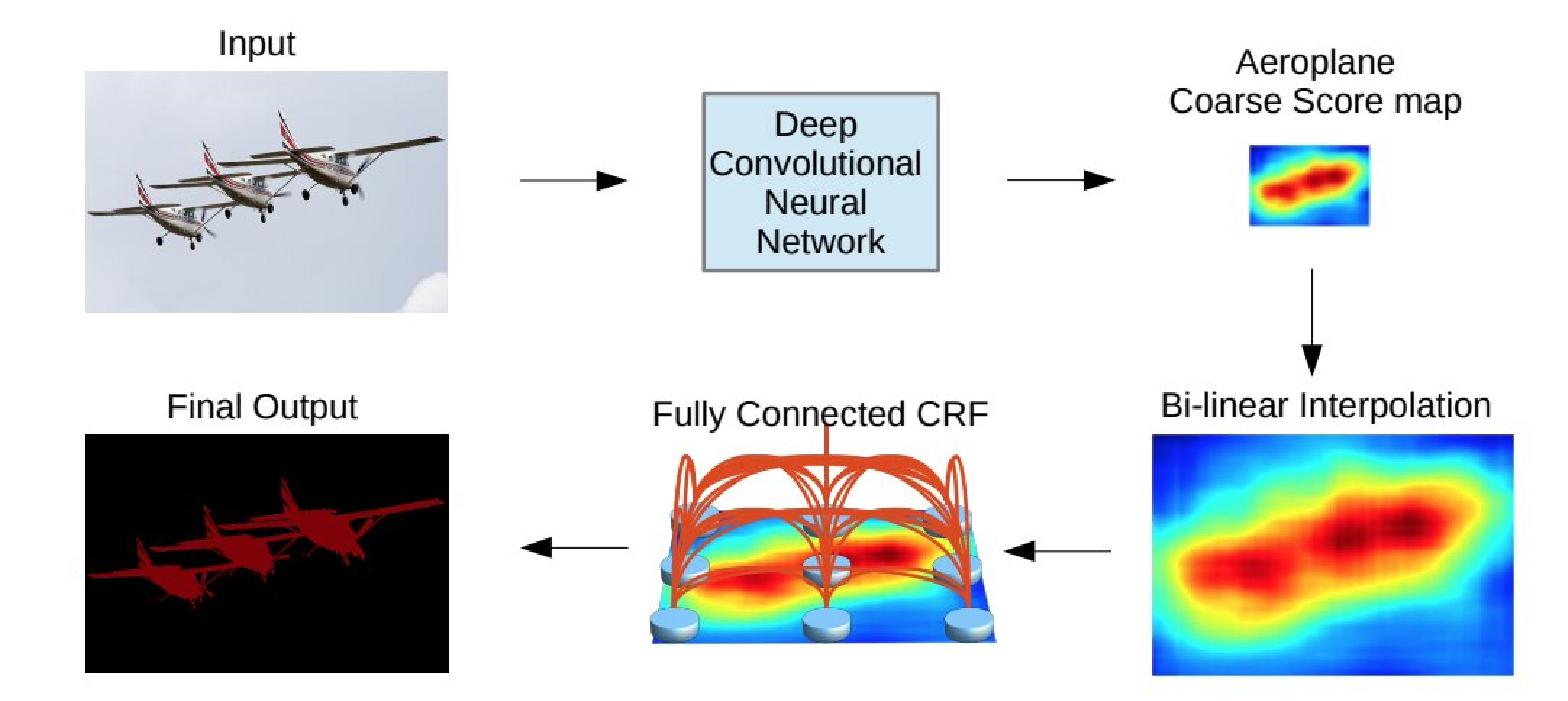}
\caption{A CNN+CRF model. The coarse score map of a CNN is up-sampled
via interpolated interpolation, and fed to a fully-connected CRF to
refine the segmentation result. From \cite{deeplab_v1}.}
\label{fig:cnn_crf}
\end{figure}

Chen \textit{et al.} \cite{deeplab_v1} proposed a semantic segmentation
algorithm based on the combination of CNNs and fully connected CRFs
(Figure~\ref{fig:cnn_crf}). They showed that responses from the final
layer of deep CNNs are not sufficiently localized for accurate object
segmentation (due to the invariance properties that make CNNs good for
high level tasks such as classification). To overcome the poor
localization property of deep CNNs, they combined the responses at the
final CNN layer with a fully-connected CRF. They showed that their
model is able to localize segment boundaries at a higher accuracy rate
than it was possible with previous methods.

Schwing and Urtasun \cite{seg_dsn} proposed a fully-connected deep
structured network for image segmentation. They presented a method
that jointly trains CNNs and fully-connected CRFs for semantic image
segmentation, and achieved encouraging results on the challenging
PASCAL VOC 2012 dataset.
In \cite{CRF-RNN}, Zheng et al. proposed a similar semantic segmentation approach integrating CRF with CNN.

In another relevant work, Lin \textit{et al.} \cite{seg_dsn2} proposed
an efficient algorithm for semantic segmentation based on contextual
deep CRFs. They explored ``patch-patch'' context (between image
regions) and ``patch-background'' context to improve semantic
segmentation through the use of contextual information.

Liu \textit{et al.} \cite{seg_dsn3} proposed a semantic segmentation
algorithm that incorporates rich information into MRFs, including
high-order relations and mixture of label contexts. Unlike previous
works that optimized MRFs using iterative algorithms, they proposed a
CNN model, namely a Parsing Network, which enables deterministic
end-to-end computation in a single forward pass.

\subsection{Encoder-Decoder Based Models}

Another popular family of deep models for image segmentation is based on the convolutional encoder-decoder architecture. 
Most of the DL-based segmentation works use some kind of encoder-decoder models.
We group these works into two categories, encoder-decoder models for general segmentation, and for medical image segmentation (to better distinguish between applications). 

\begin{figure}[h]
\centering
\includegraphics[width=0.9\linewidth]{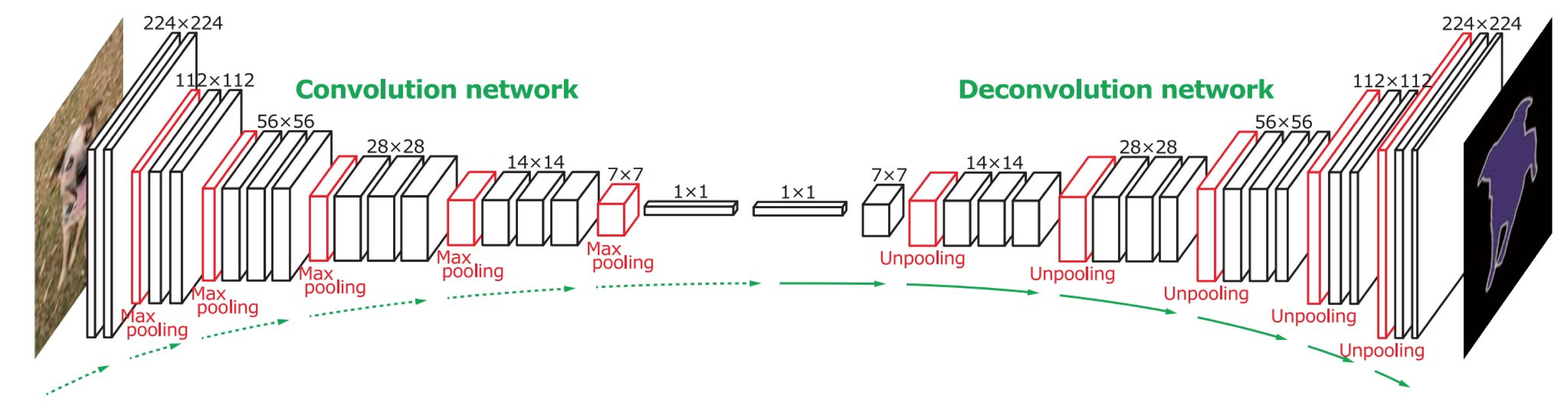}
\caption{Deconvolutional semantic segmentation. Following a
convolution network based on the VGG 16-layer net, is a multi-layer
deconvolution network to generate the accurate segmentation map. From
\cite{seg_decon1}.}
\label{fig:seg_decon}
\end{figure}

\textbf{\subsubsection{Encoder-Decoder Models for General Segmentation}}
Noh \textit{et al.} \cite{seg_decon1} published an early paper on
semantic segmentation based on deconvolution (a.k.a.~transposed
convolution). Their model (Figure~\ref{fig:seg_decon}) consists of two parts, an encoder using convolutional layers adopted from the VGG
16-layer network and a deconvolutional network that takes the feature vector as input and generates a map of pixel-wise class probabilities. The deconvolution network is composed of deconvolution
and unpooling layers, which identify pixel-wise class labels and
predict segmentation masks.

This network achieved promising performance on the PASCAL VOC 2012
dataset, and obtained the best accuracy (72.5\%) among the methods
trained with no external data at the time.

\begin{figure}[h]
\centering
\includegraphics[width=0.8\linewidth]{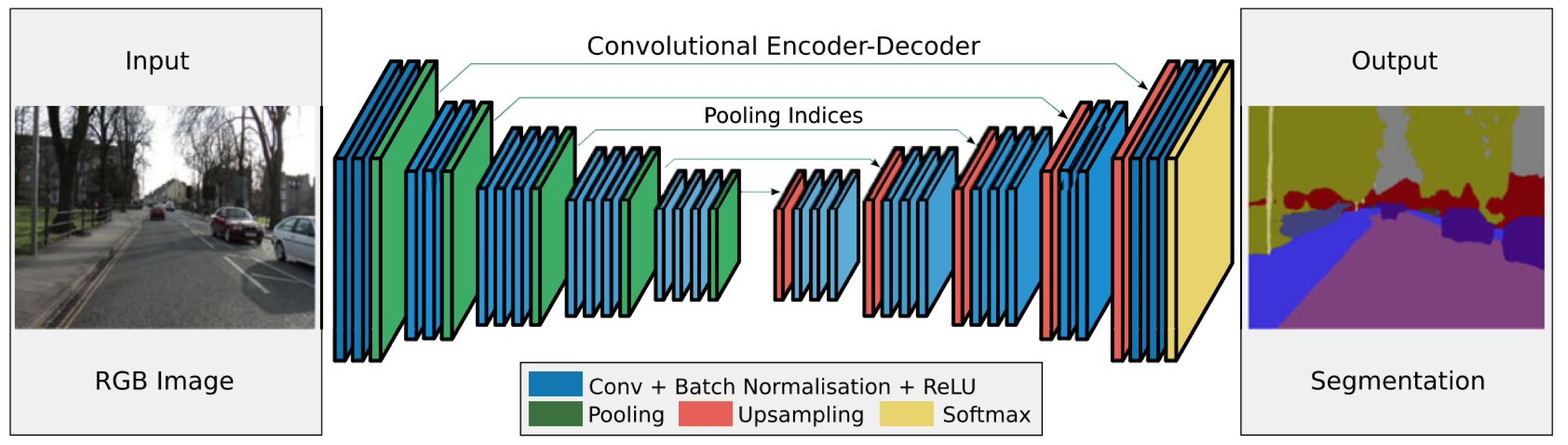}
\caption{SegNet has no fully-connected layers; hence, the model is
fully convolutional. A decoder up-samples its input using the
transferred pool indices from its encoder to produce a sparse feature
map(s). From \cite{segnet}.}
\label{fig:segnet_arc}
\end{figure}

In another promising work known as SegNet, Badrinarayanan \textit{et
al.} \cite{segnet} proposed a convolutional encoder-decoder
architecture for image segmentation (Figure~\ref{fig:segnet_arc}).
Similar to the deconvolution network, the core trainable segmentation
engine of SegNet consists of an encoder network, which is
topologically identical to the 13 convolutional layers in the VGG16
network, and a corresponding decoder network followed by a pixel-wise
classification layer. The main novelty of SegNet is in the way the
decoder upsamples its lower resolution input feature map(s);
specifically, it uses pooling indices computed in the max-pooling step
of the corresponding encoder to perform non-linear up-sampling. This
eliminates the need for learning to up-sample. The (sparse) up-sampled
maps are then convolved with trainable filters to produce dense
feature maps. SegNet is also significantly smaller in the number of
trainable parameters than other competing architectures. A Bayesian
version of SegNet was also proposed by the same authors to model the
uncertainty inherent to the convolutional encoder-decoder network for
scene segmentation \cite{bayes_segnet}.

\begin{figure*}
\centering
\includegraphics[width=0.9\linewidth]{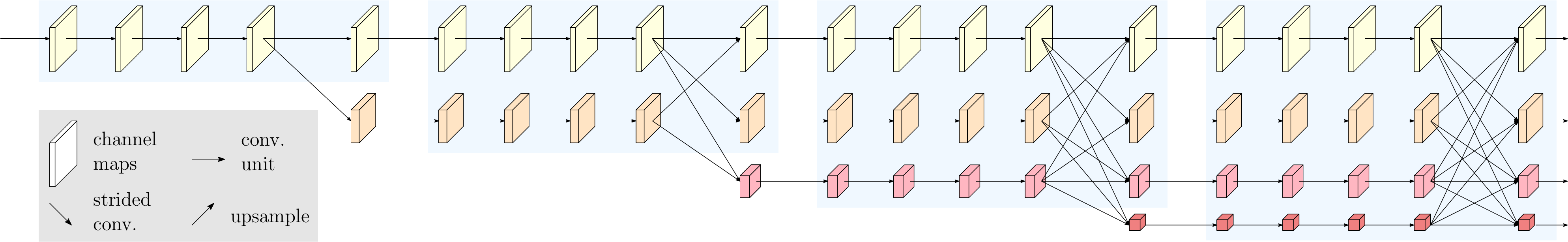}
\caption{Illustrating the HRNet architecture. It consists of parallel high-to-low resolution convolution streams with repeated information exchange across multi-resolution steams. There are four stages. The 1st stage consists of high-resolution convolutions. The 2nd (3rd, 4th) stage repeats two-resolution (three-resolution, four-resolution) blocks. From \cite{hrrocr}.}
\label{fig:HRNet}
\end{figure*}

Another popular model in this category is the recently-developed segmentation network, high-resolution network (HRNet)~\cite{hrrocr} Figure~\ref{fig:HRNet}. Other than recovering high-resolution representations as done in DeConvNet, SegNet, U-Net and V-Net, HRNet maintains high-resolution representations through the encoding process by connecting the high-to-low resolution convolution streams in parallel, and  repeatedly exchanging the information across resolutions. Many of the more recent works on semantic segmentation use HRNet as the backbone by exploiting contextual models, such as self-attention and its extensions.

Several other works adopt transposed convolutions, or encoder-decoders for image segmentation, such as Stacked Deconvolutional Network (SDN) \cite{SDN}, Linknet \cite{linknet}, W-Net \cite{wnet}, and locality-sensitive deconvolution networks for RGB-D segmentation \cite{cheng2017locality}.
One limitation of Encoder-Decoder based models is the loss of fine-grained information of the image, due to the loss of high-resolution representations through the encoding process. This issue is however addressed in some of the recent architectures such as HR-Net.


\textbf{\subsubsection{Encoder-Decoder Models for Medical and Biomedical Image Segmentation}}

There are several models initially developed for medical/biomedical image segmentation, which are inspired by FCNs and encoder-decoder models. 
U-Net \cite{unet}, and V-Net \cite{vnet}, are two well-known such architectures, which are now also being used outside the medical domain.

\begin{figure}[h]
\centering
\includegraphics[width=0.8\linewidth]{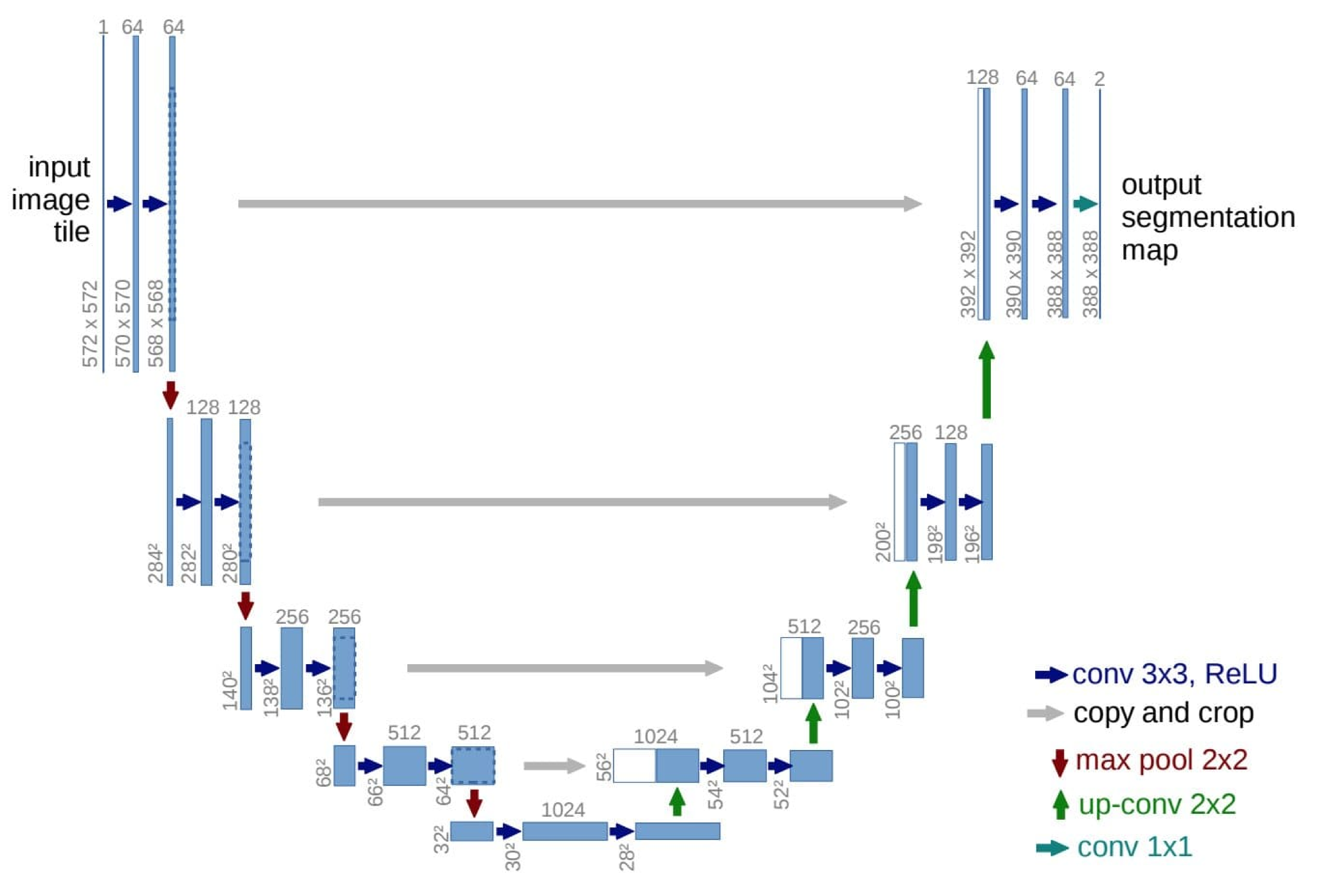}
\caption{The U-net model. The blue boxes denote feature map blocks
with their indicated shapes. From \cite{unet}.}
\label{fig:unet}
\end{figure}

Ronneberger \textit{et al.} \cite{unet} proposed the U-Net for
segmenting biological microscopy images. Their network and training
strategy relies on the use of data augmentation to learn from the very few annotated images effectively. The U-Net architecture
(Figure~\ref{fig:unet}) comprises two parts, a contracting path to
capture context, and a symmetric expanding path that enables precise
localization. The down-sampling or contracting part has a FCN-like
architecture that extracts features with $3\times3$ convolutions. The
up-sampling or expanding part uses up-convolution (or deconvolution),
reducing the number of feature maps while increasing their dimensions.
Feature maps from the down-sampling part of the network are copied to
the up-sampling part to avoid losing pattern information. Finally, a
$1\times1$ convolution processes the feature maps to generate a
segmentation map that categorizes each pixel of the input image. 
U-Net was trained on 30 transmitted light microscopy images, and it won the ISBI cell tracking challenge 2015 by
a large margin.

Various extensions of U-Net have been developed for different kinds of
images. For example, Cicek \cite{unet3d} proposed a U-Net architecture
for 3D images. Zhou \textit{et al.} \cite{unetplus} developed a nested
U-Net architecture. 
U-Net has also been applied to various other problems. For example,
Zhang \textit{et al.} \cite{unet_road} developed a road
segmentation/extraction algorithm based on U-Net.

V-Net is another well-known, FCN-based model,
which was proposed by Milletari \textit{et al.} \cite{vnet} for 3D
medical image segmentation. For model training, they introduced a new
objective function based on Dice coefficient, enabling the model
to deal with situations in which there is a strong imbalance between
the number of voxels in the foreground and background. The network was
trained end-to-end on MRI volumes of prostate, and learns to
predict segmentation for the whole volume at once.
Some of the other relevant works on medical image segmentation includes Progressive Dense V-net (PDV-Net) \textit{et al.} for fast and automatic segmentation of pulmonary lobes from chest CT images, and the 3D-CNN encoder for lesion segmentation \cite{brosch2016deep}.

\subsection{Multi-Scale and Pyramid Network Based Models}
Multi-scale analysis, a rather old idea in image processing, has been
deployed in various neural network architectures.
One of the most prominent models of this sort is the Feature Pyramid
Network (FPN) proposed by Lin \textit{et al.} \cite{fpn}, which was
developed mainly for object detection but was then also applied to
segmentation. The inherent multi-scale, pyramidal hierarchy of deep
CNNs was used to construct feature pyramids with marginal extra cost.
To merge low and high resolution features, the FPN is composed of a
bottom-up pathway, a top-down pathway and lateral connections. The
concatenated feature maps are then processed by a $3\times3$
convolution to produce the output of each stage. Finally, each stage
of the top-down pathway generates a prediction to detect an object.
For image segmentation, the authors use two multi-layer perceptrons
(MLPs) to generate the masks.

\begin{figure}
\centering
\includegraphics[width=0.9\linewidth]{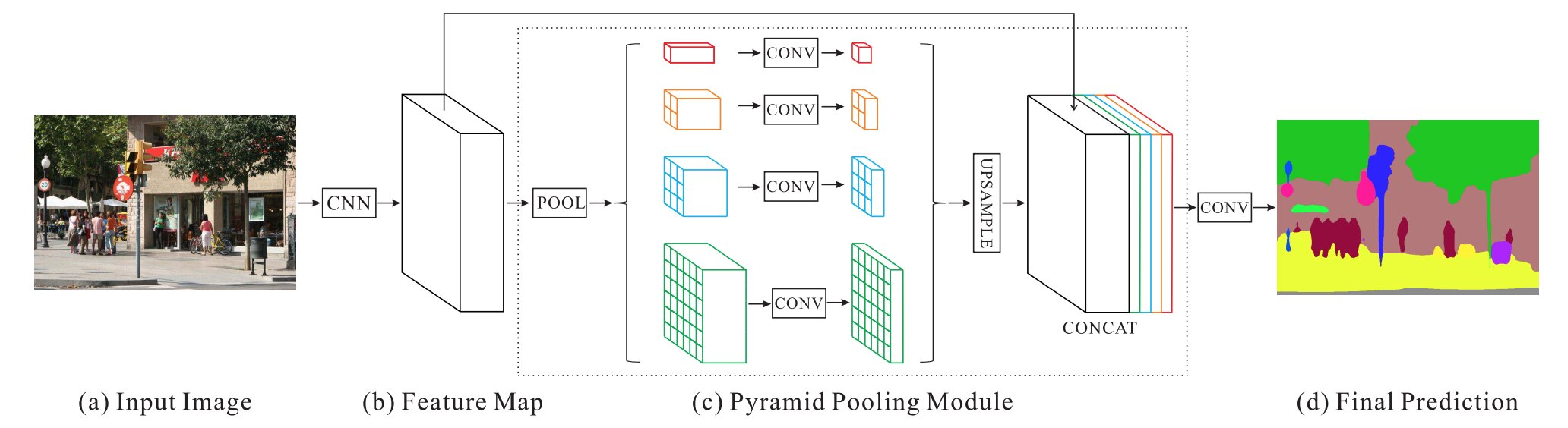}
\caption{The PSPN architecture. A CNN produces the feature map and a
pyramid pooling module aggregates the different sub-region
representations. Up-sampling and concatenation are used to form the
final feature representation from which, the final pixel-wise
prediction is obtained through convolution. From \cite{pspn}.}
\label{fig:pspn}
\end{figure}

Zhao \textit{et al.} \cite{pspn} developed the Pyramid Scene Parsing
Network (PSPN), a multi-scale network to better learn the global
context representation of a scene (Figure~\ref{fig:pspn}). Different
patterns are extracted from the input image using a residual network
(ResNet) as a feature extractor, with a dilated network. These feature
maps are then fed into a pyramid pooling module to distinguish
patterns of different scales. They are pooled at four different
scales, each one corresponding to a pyramid level and processed by a
$1\times1$ convolutional layer to reduce their dimensions. The outputs
of the pyramid levels are up-sampled and concatenated with the initial
feature maps to capture both local and global context information.
Finally, a convolutional layer is used to generate the pixel-wise
predictions. 

Ghiasi and Fowlkes \cite{laplace} developed a multi-resolution reconstruction architecture based on a
Laplacian pyramid that uses skip connections from higher resolution
feature maps and multiplicative gating to successively refine segment
boundaries reconstructed from lower-resolution maps. They showed that,
while the apparent spatial resolution of convolutional feature maps is
low, the high-dimensional feature representation contains significant
sub-pixel localization information.

There are other models using multi-scale analysis for segmentation, such as DM-Net (Dynamic Multi-scale Filters Network) \cite{dmsf}, Context contrasted network and gated multi-scale aggregation (CCN) \cite{CCL}, Adaptive Pyramid Context Network (APC-Net) \cite{apcnet}, Multi-scale context intertwining (MSCI) \cite{MSCI}, 
and salient object segmentation
\cite{salient_seg}.

\subsection{R-CNN Based Models (for Instance Segmentation)}

\begin{figure}[h]
\centering
\includegraphics[width=0.4\linewidth]{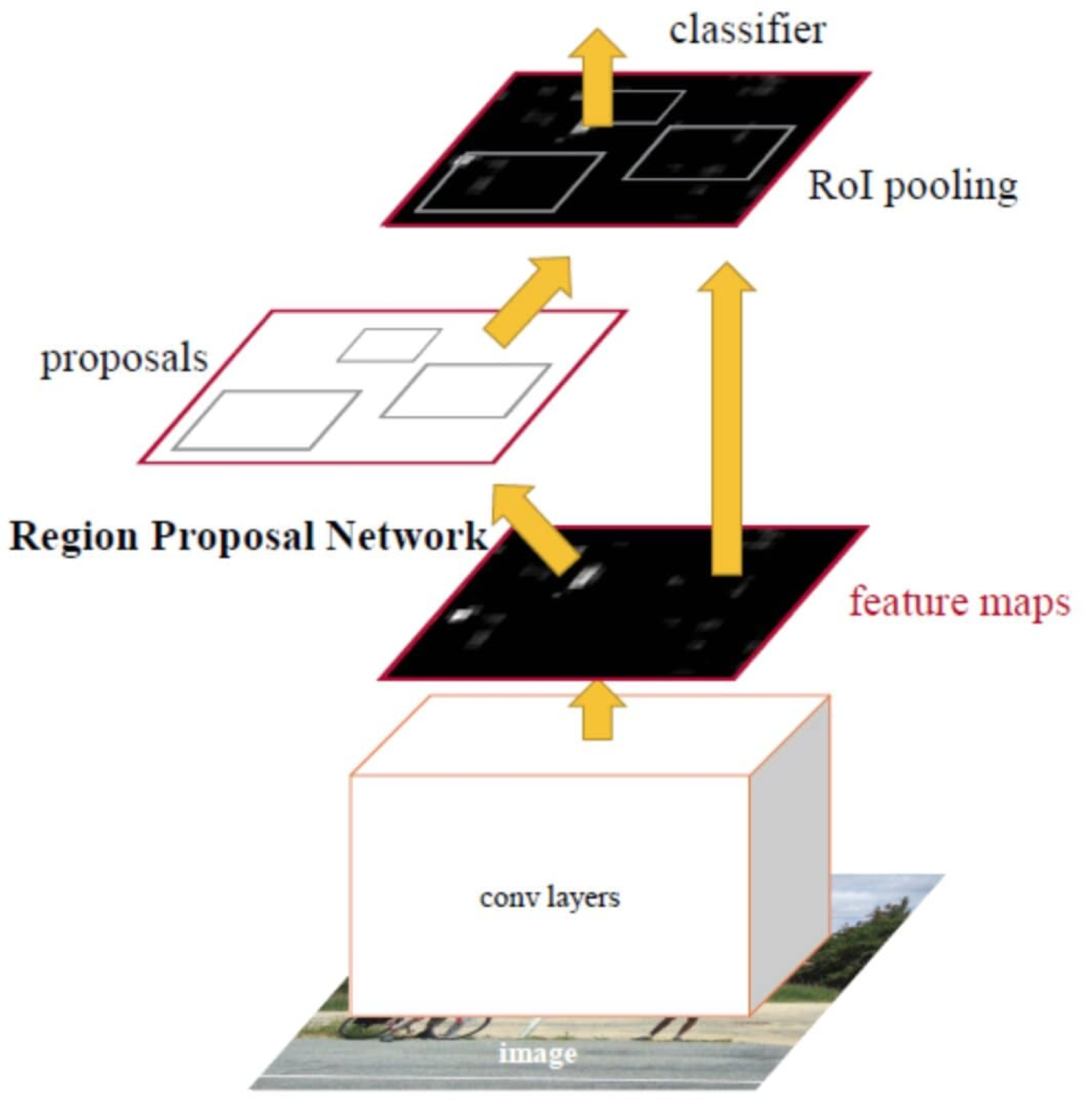}
\caption{Faster R-CNN architecture. 
Courtesy of \cite{faster_rcnn}.}
\label{fig:faster_rcnn}
\end{figure}

The regional convolutional network (R-CNN) and its extensions (Fast
R-CNN, Faster R-CNN, Maksed-RCNN) have proven successful in object detection
applications. 
In particular, the Faster R-CNN \cite{faster_rcnn} architecture
(Figure~\ref{fig:faster_rcnn}) developed for object detection uses a
region proposal network (RPN) to propose bounding box candidates. The
RPN extracts a Region of Interest (RoI), and a RoIPool layer computes
features from these proposals in order to infer the bounding box coordinates and the class of the object.
Some of the extensions of R-CNN have been heavily used to address the
instance segmentation problem; i.e., the task of simultaneously
performing object detection and semantic segmentation. 

\begin{figure}[h]
\centering
\includegraphics[width=0.6\linewidth]{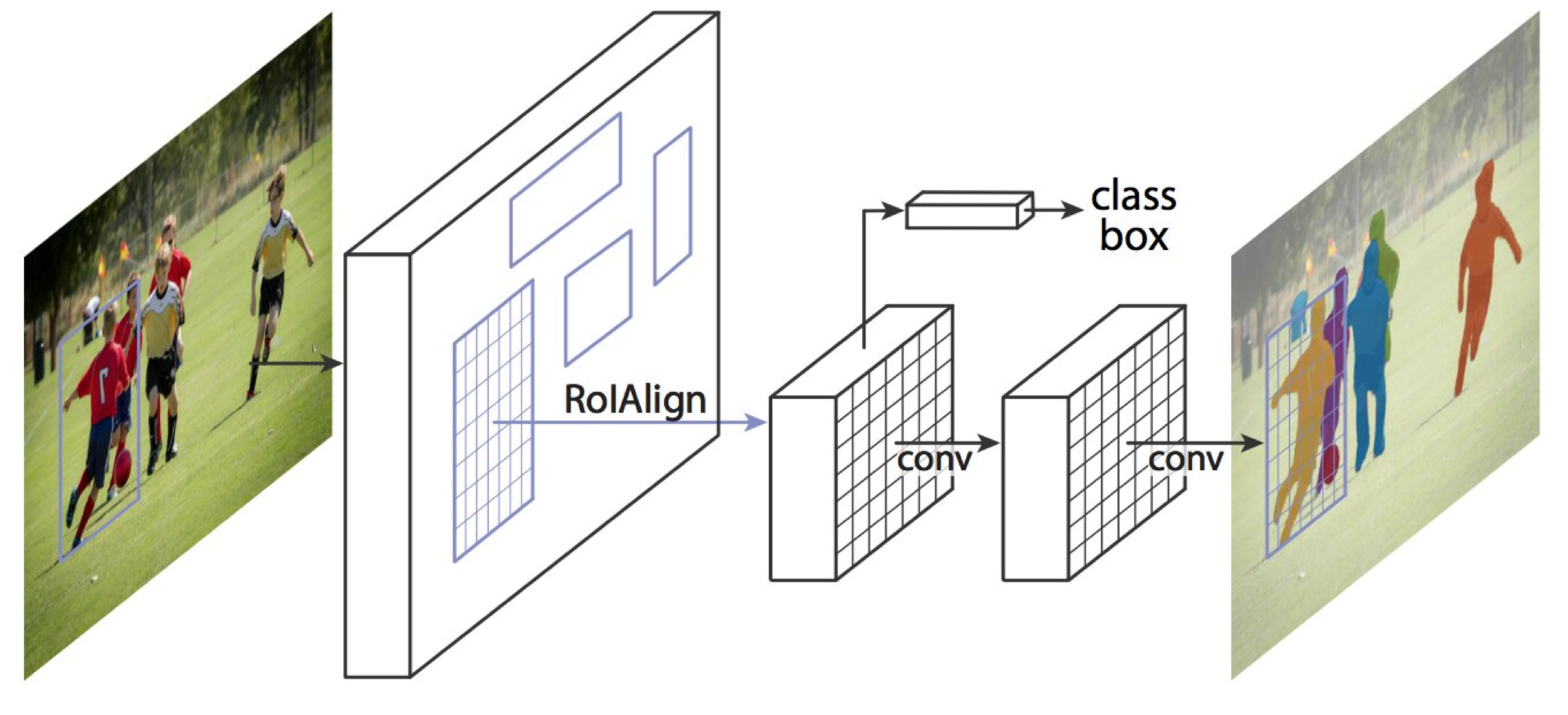}
\caption{Mask R-CNN architecture for instance segmentation. From
\cite{mask_rcnn}.}
\label{fig:mask_rcnn}
\end{figure}

\begin{figure}[h]
\centering
\includegraphics[width=0.7\linewidth]{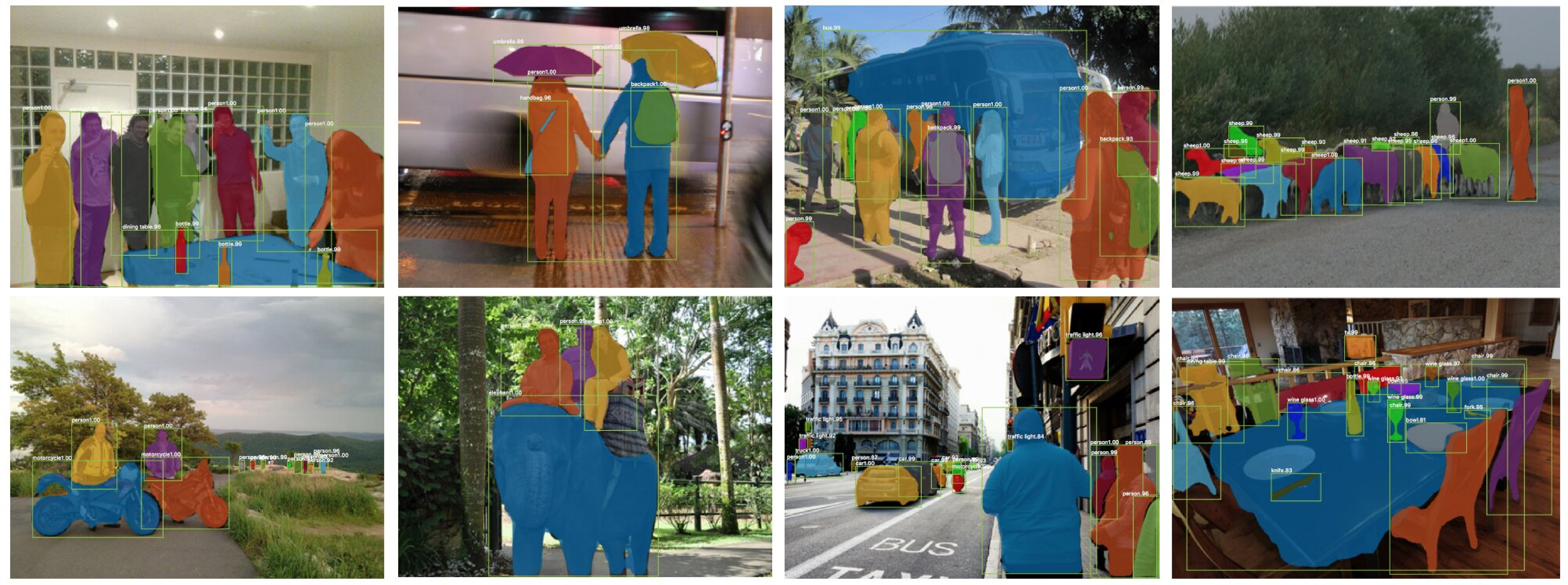}
\caption{Mask R-CNN results on sample images from the COCO test set.
From \cite{mask_rcnn}.}
\label{fig:mask_rcnn_Res}
\end{figure}

In one extension of this model, He \textit{et al.} \cite{mask_rcnn}
proposed a Mask R-CNN for object instance segmentation, which beat all
previous benchmarks on many COCO challenges. 
This model efficiently
detects objects in an image while simultaneously generating a
high-quality segmentation mask for each instance. Mask R-CNN is
essentially a Faster R-CNN with 3 output branches
(Figure~\ref{fig:mask_rcnn})---the first computes the bounding box
coordinates, the second computes the associated classes, and the third
computes the binary mask to segment the object.
The Mask R-CNN loss function combines the losses of the bounding box
coordinates, the predicted class, and the segmentation mask, and
trains all of them jointly. 
Figure~\ref{fig:mask_rcnn_Res} shows the Mask-RCNN result on some sample images.

\begin{figure}[h]
\centering
\includegraphics[width=0.7\linewidth]{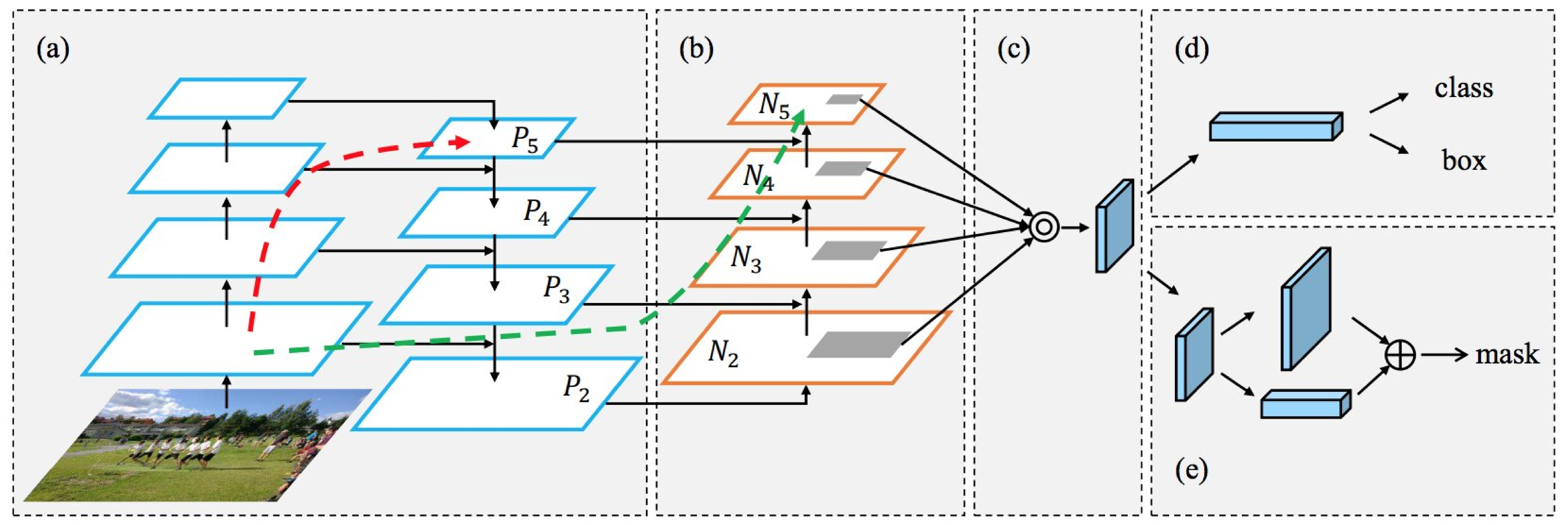}
\caption{The Path Aggregation Network. (a) FPN backbone. (b) Bottom-up
path augmentation. (c) Adaptive feature pooling. (d) Box branch. (e)
Fully-connected fusion. Courtesy of \cite{seg_pan}.}
\label{fig:seg_pan}
\end{figure}

The Path Aggregation Network (PANet) proposed by Liu \textit{et al.}
\cite{seg_pan} is based on the Mask R-CNN and FPN models
(Figure~\ref{fig:seg_pan}). The feature extractor of the network uses
an FPN architecture with a new augmented bottom-up pathway improving
the propagation of low-layer features. Each stage of this third
pathway takes as input the feature maps of the previous stage and
processes them with a $3\times3$ convolutional layer. The output is
added to the same stage feature maps of the top-down pathway using a
lateral connection and these feature maps feed the next stage.
As in the Mask R-CNN, the output of the adaptive feature pooling layer
feeds three branches. The first two use a fully connected layer to
generate the predictions of the bounding box coordinates and the
associated object class. The third processes the RoI with an FCN to
predict the object mask.

Dai \textit{et al.} \cite{seg_Cascades} developed a multi-task network
for instance-aware semantic segmentation, that consists of three
networks, respectively differentiating instances, estimating masks,
and categorizing objects. These networks form a cascaded structure,
and are designed to share their convolutional features.
Hu \textit{et al.} \cite{seg_learn} proposed a new partially-supervised training paradigm, together with a novel weight
transfer function, that enables training instance segmentation models
on a large set of categories, all of which have box annotations, but
only a small fraction of which have mask annotations.

\begin{figure}[h]
\centering
\includegraphics[width=0.9\linewidth]{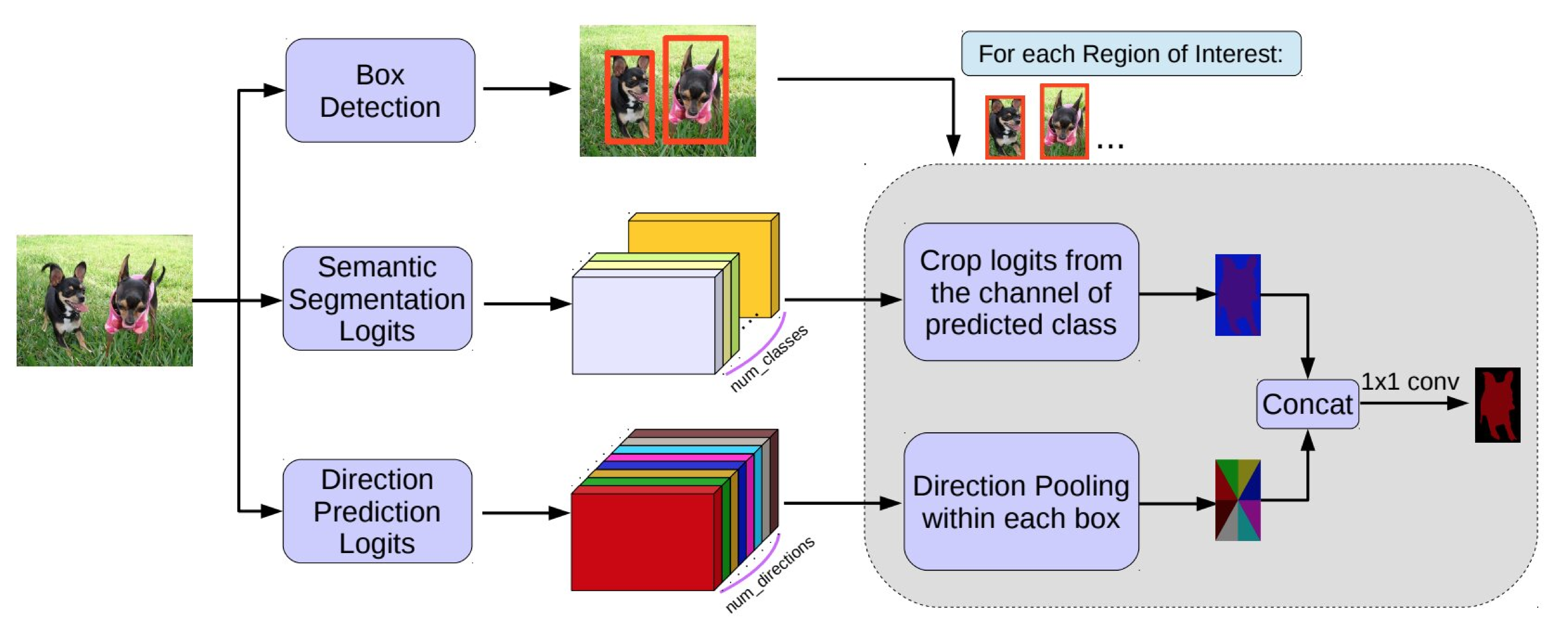}
\caption{The MaskLab model. MaskLab generates three outputs---refined
box predictions (from Faster R-CNN), semantic segmentation logits for
pixel-wise classification, and direction prediction logits for
predicting each pixel's direction toward its instance center. From
\cite{masklab}.}
\label{fig:masklab}
\end{figure}

Chen \textit{et al.} \cite{masklab} developed an instance segmentation
model, MaskLab (Figure~\ref{fig:masklab}), by refining object
detection with semantic and direction features based on Faster R-CNN.
This model produces three outputs, box detection, semantic
segmentation, and direction prediction. Building on the Faster-RCNN
object detector, the predicted boxes provide accurate localization of
object instances. Within each region of interest, MaskLab performs
foreground/background segmentation by combining semantic and direction
prediction.

Another interesting model is Tensormask, proposed by Chen \textit{et
al.} \cite{TensorMask}, which is based on dense sliding window
instance segmentation. They treat dense instance segmentation as a
prediction task over 4D tensors and present a general framework that
enables novel operators on 4D tensors. They demonstrate that the
tensor view leads to large gains over baselines and yields results
comparable to Mask R-CNN. TensorMask achieves promising results on
dense object segmentation.

Many other instance segmentation models have been developed based on
R-CNN, such as those developed for mask proposals, including R-FCN
\cite{RFCN}, 
DeepMask \cite{DeepMask}, PolarMask \cite{polarmask}, boundary-aware
instance segmentation \cite{Hayer}, and CenterMask \cite{lee2020centermask}.
It is worth noting that there is another promising research direction that attempts to solve the instance segmentation problem by learning grouping cues for bottom-up segmentation, such as Deep Watershed Transform \cite{bai2017deep}, real-time instance segmentation \cite{bolya2019yolact}, and Semantic Instance Segmentation via Deep Metric Learning \cite{fathi2017semantic}.

\subsection{Dilated Convolutional Models and DeepLab Family}

\begin{figure}[h]
\centering
\includegraphics[width=0.7\linewidth]{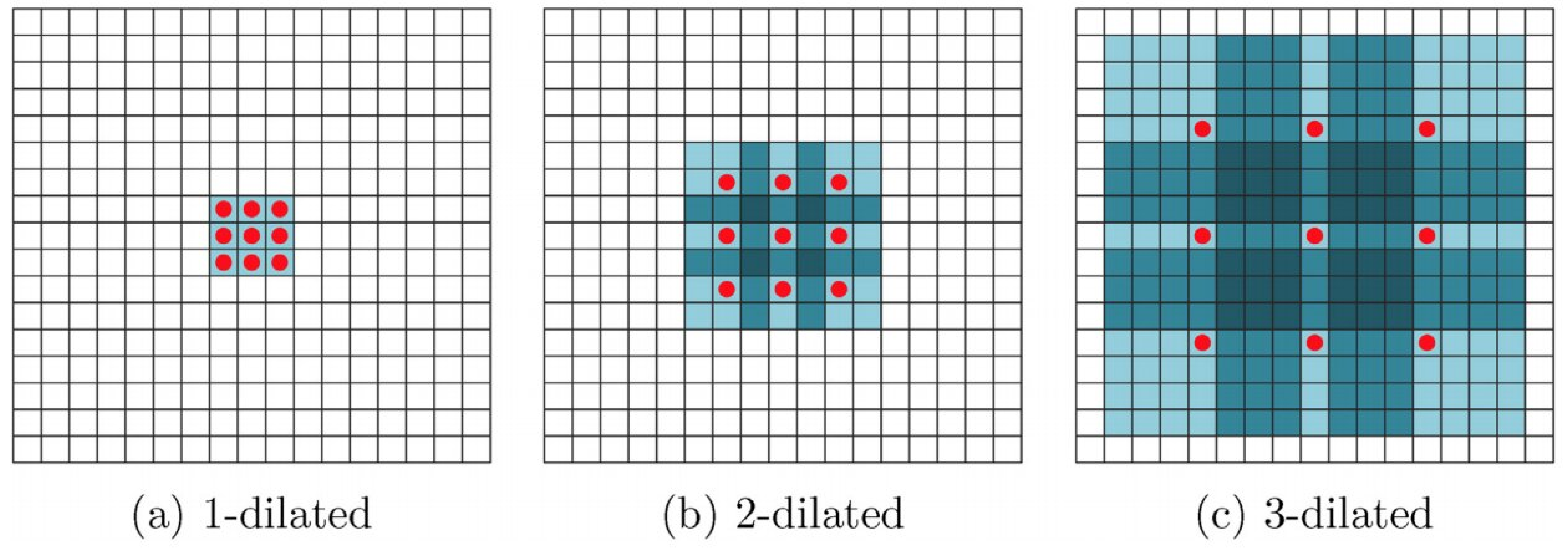}
\caption{Dilated convolution. A $3\times3$ kernel at different
dilation rates.}
\label{fig_dilation}
\end{figure}

Dilated convolution (a.k.a. ``atrous'' convolution) introduces another
parameter to convolutional layers, the dilation rate. The dilated
convolution (Figure~\ref{fig_dilation}) of a signal $x(i)$ is defined
as $y_i= \sum_{k=1}^K x[i+rk] w[k]$, where $r$ is the dilation rate
that defines a spacing between the weights of the kernel $w$. For
example, a $3\times3$ kernel with a dilation rate of 2 will have the
same size receptive field as a $5\times5$ kernel while using only 9
parameters, thus enlarging the receptive field with no increase in
computational cost. Dilated convolutions have been popular in the
field of real-time segmentation, and many recent publications report
the use of this technique. Some of most important include the DeepLab
family \cite{deeplab}, multi-scale context aggregation
\cite{multi_cont_agg},  dense  upsampling  convolution and hybrid dilatedconvolution (DUC-HDC) \cite{UCS}, densely connected Atrous Spatial Pyramid Pooling (DenseASPP) \cite{Denseaspp}, and the efficient neural network (ENet) \cite{Enet}.

\begin{figure}[h]
\centering
\includegraphics[width=0.7\linewidth]{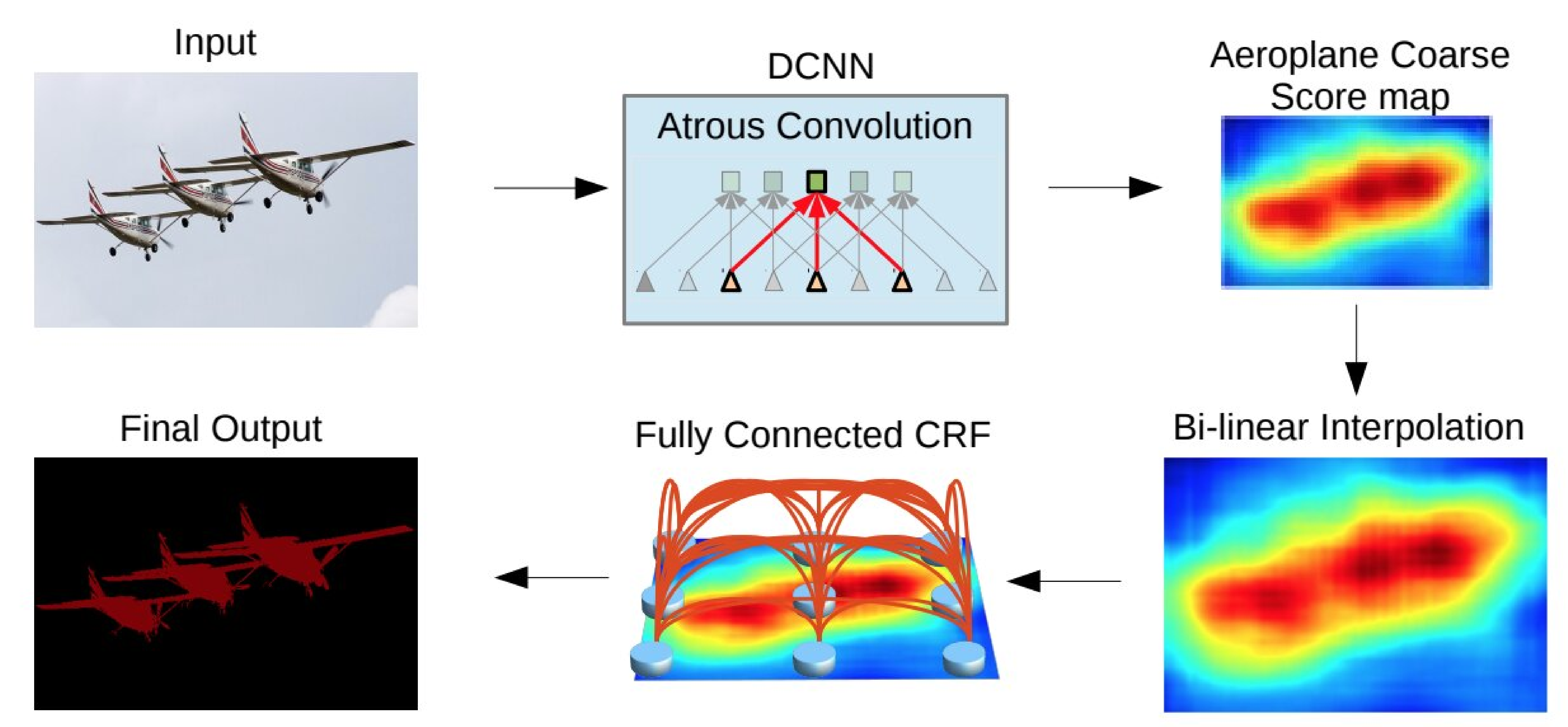}
\caption{The DeepLab model. A CNN model such as VGG-16 or ResNet-101
is employed in fully convolutional fashion, using dilated convolution.
A bilinear interpolation stage enlarges the feature maps to the
original image resolution. Finally, a fully connected CRF refines the
segmentation result to better capture the object boundaries. From
\cite{deeplab}}
\label{deeplab}
\end{figure}

DeepLabv1 \cite{deeplab_v1} and DeepLabv2
\cite{deeplab} are among some of the most popular image segmentation approaches, developed by Chen \textit{et al.}. 
The latter has three key features. First is the use of dilated convolution
to address the decreasing resolution in the network (caused by
max-pooling and striding). Second is Atrous Spatial Pyramid Pooling (ASPP), which probes an incoming
convolutional feature layer with filters at multiple sampling rates, thus capturing objects as well
as image context at multiple scales to robustly segment objects at
multiple scales. Third is improved localization of object boundaries
by combining methods from deep CNNs and probabilistic graphical
models. 
The best DeepLab (using a ResNet-101 as backbone) has reached a
79.7\% mIoU score on the 2012 PASCAL VOC challenge, a 45.7\% mIoU
score on the PASCAL-Context challenge and a 70.4\% mIoU score on the
Cityscapes challenge. Figure~\ref{deeplab} illustrates the Deeplab model, which is similar to \cite{deeplab_v1}, the main difference being the use of dilated convolution and ASPP.

Subsequently, Chen \textit{et al.} \cite{deeplabv3} proposed
DeepLabv3, which combines cascaded and parallel modules of dilated
convolutions. The parallel convolution modules are grouped in the
ASPP. A $1\times1$ convolution and batch normalisation are added in
the ASPP. All the outputs are concatenated and processed by another
$1\times1$ convolution to create the final output with logits for each
pixel.

\begin{figure}[h]
\centering
\includegraphics[width=0.9\linewidth]{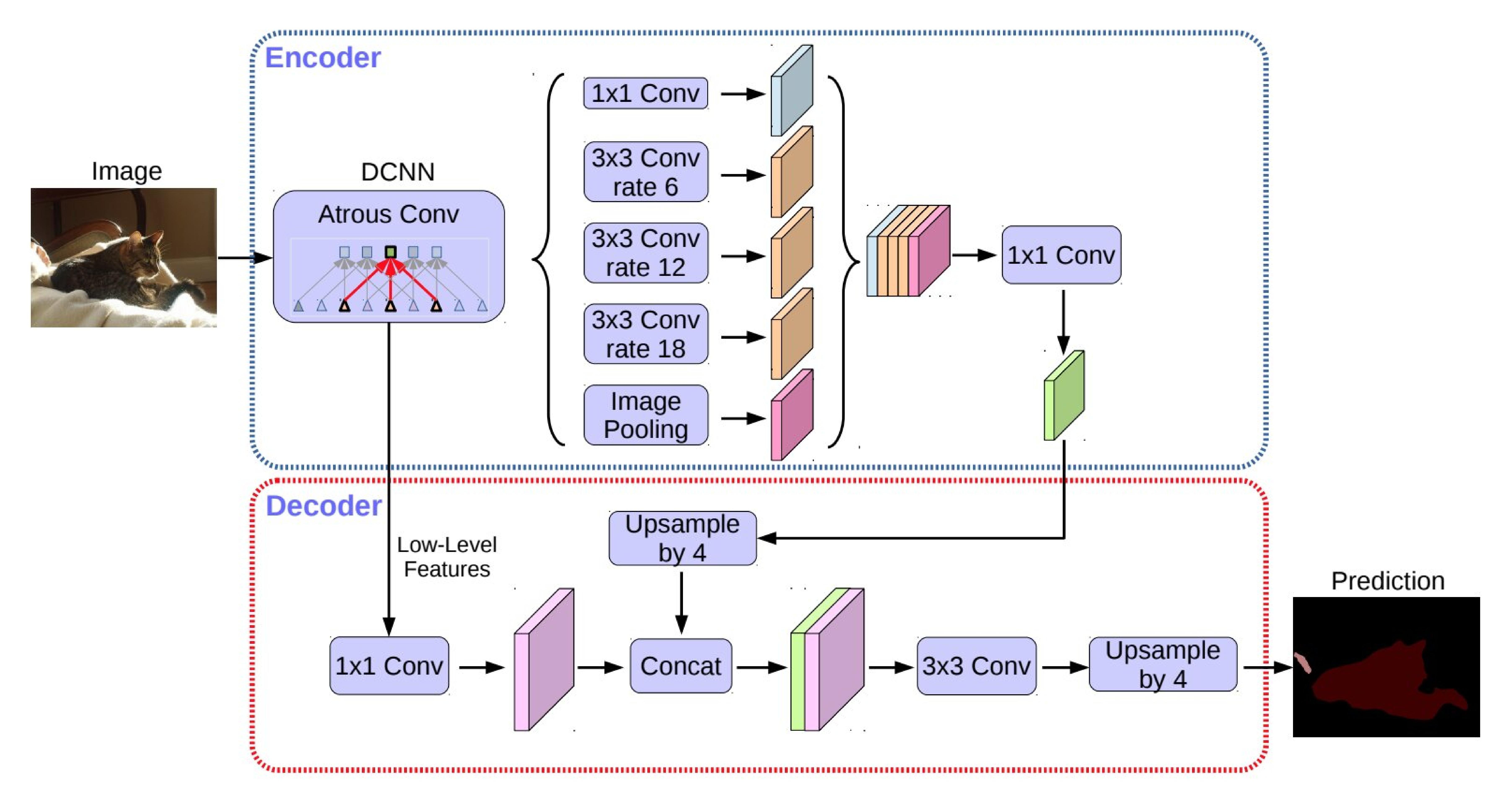}
\caption{The DeepLabv3+ model. From \cite{deeplabv3plus}.}
\label{deeplabplus}
\end{figure}

In 2018, Chen \textit{et al.} \cite{deeplabv3plus} released
Deeplabv3+, which uses an encoder-decoder architecture
(Figure~\ref{deeplabplus}), including atrous separable convolution,
composed of a depthwise convolution (spatial convolution for each
channel of the input) and pointwise convolution ($1\times1$
convolution with the depthwise convolution as input). They used the
DeepLabv3 framework as encoder. The most relevant model has a modified
Xception backbone with more layers, dilated depthwise separable
convolutions instead of max pooling and batch normalization. The best
DeepLabv3+ pretrained on the COCO and the JFT datasets has obtained a
89.0\% mIoU score on the 2012 PASCAL VOC challenge.

\subsection{Recurrent Neural Network Based Models}
\label{sec:RNN}

While CNNs are a natural fit for computer vision problems, they are
not the only possibility. RNNs are useful in modeling the short/long
term dependencies among pixels to (potentially) improve the estimation
of the segmentation map. Using RNNs, pixels may be linked together and
processed sequentially to model global contexts and improve semantic
segmentation. One challenge, though, is the natural 2D structure of
images.

\begin{figure}[h]
\centering
\includegraphics[width=0.9\linewidth]{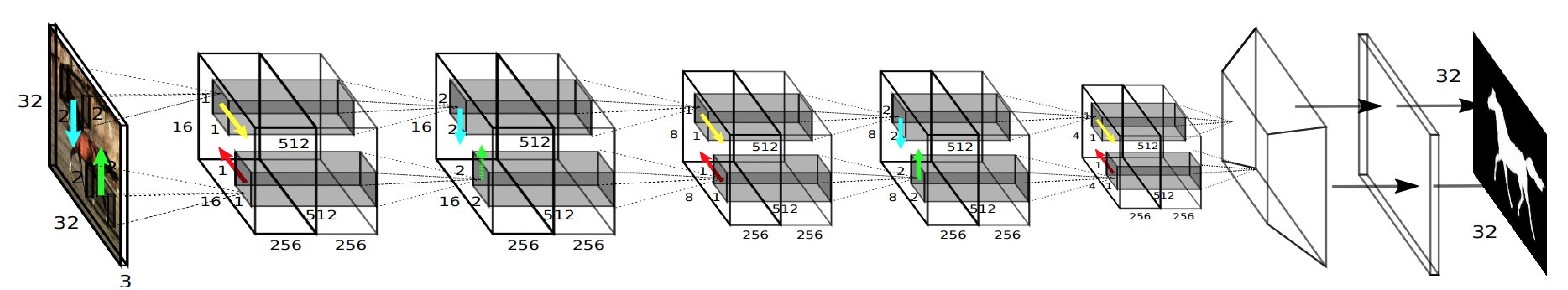}
\caption{The ReSeg model. The pre-trained VGG-16 feature extractor
network is not shown. From \cite{Reseg}.}
\label{fig:reseg}
\end{figure}

Visin \textit{et al.} \cite{Reseg} proposed an RNN-based model for
semantic segmentation called ReSeg. This model is mainly based on
another work, ReNet \cite{renet}, which was developed for image
classification.
Each ReNet layer 
is composed of four RNNs
that sweep the image horizontally and vertically in both directions,
encoding patches/activations, and providing relevant global
information. To perform image segmentation with the ReSeg model
(Figure~\ref{fig:reseg}), ReNet layers are stacked on top of
pre-trained VGG-16 convolutional layers that extract generic local
features. ReNet layers are then followed by up-sampling layers to
recover the original image resolution in the final predictions. Gated
Recurrent Units (GRUs) are used because they provide a good balance
between memory usage and computational power.

In another work, Byeon \textit{et al.} \cite{seg_lstm} developed a
pixel-level segmentation and classification of scene images using
long-short-term-memory (LSTM) network. They investigated
two-dimensional (2D) LSTM networks for images of natural scenes,
taking into account the complex spatial dependencies of labels. In
this work, classification, segmentation, and context integration are
all carried out by 2D LSTM networks, allowing texture and spatial
model parameters to be learned within a single model. 

\begin{figure}[h]
\centering
\includegraphics[width=0.7\linewidth]{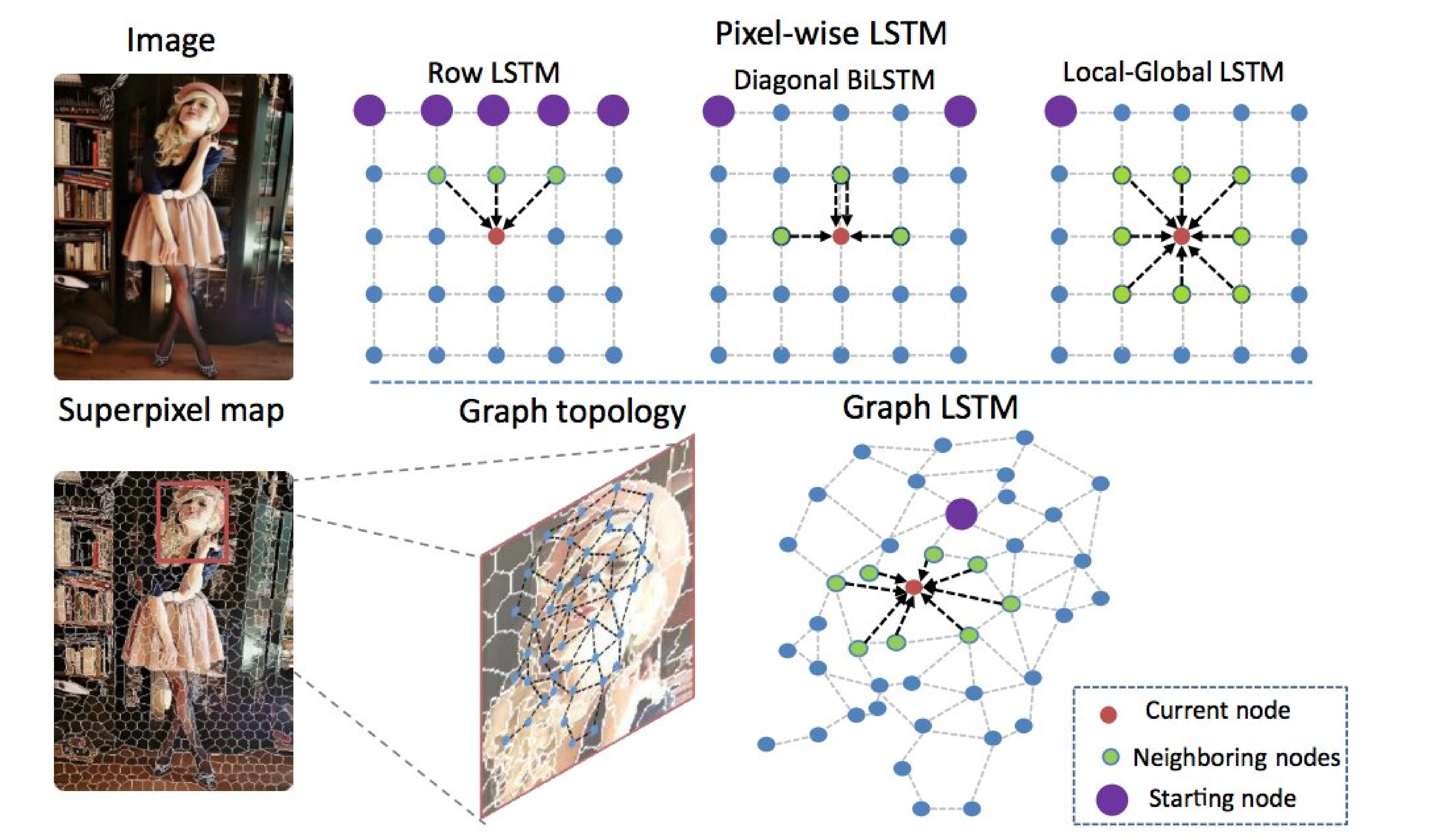}
\caption{Comparison between the graph-LSTM model and traditional
pixel-wise RNN models. From \cite{graph_lstm}.}
\label{fig:graph_lstm}
\end{figure}

\begin{figure}[h]
\centering
\includegraphics[width=0.8\linewidth]{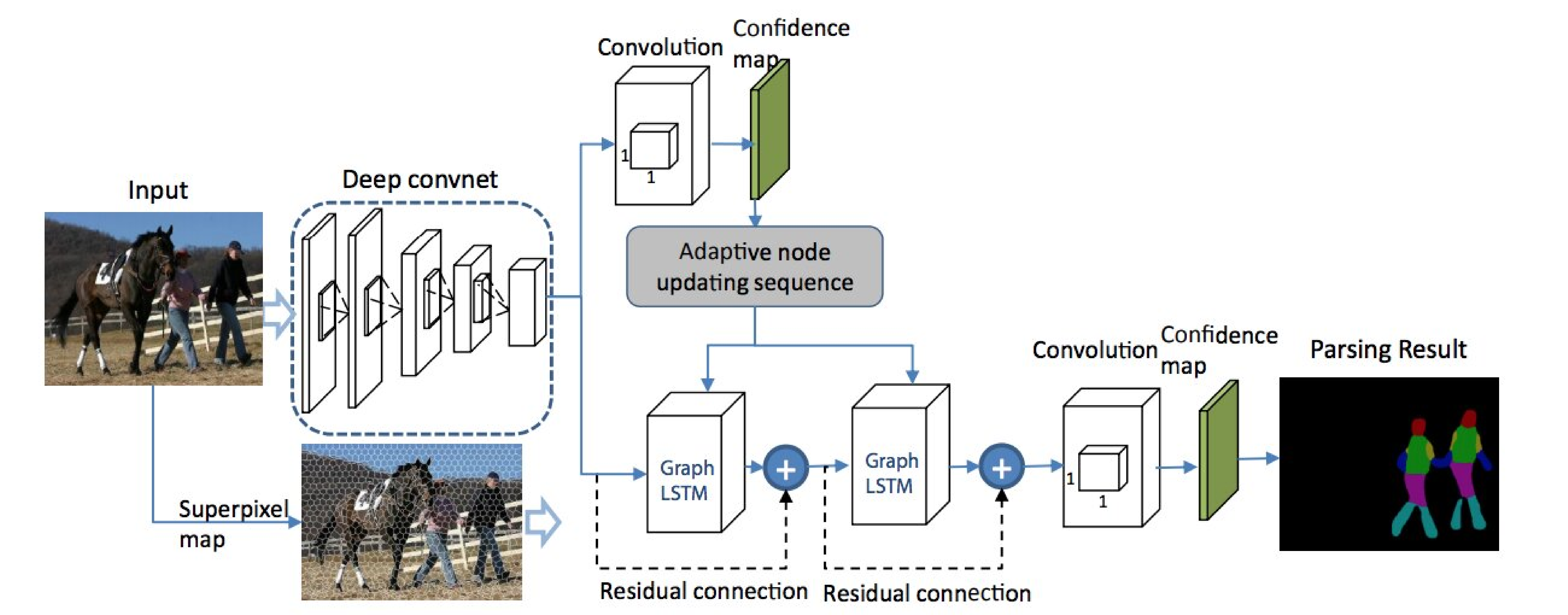}
\caption{The graph-LSTM model for semantic segmentation. From
\cite{graph_lstm}.}
\label{fig:seg_graph_lstm}
\end{figure}

Liang \textit{et al.} \cite{graph_lstm} proposed a semantic
segmentation model based on the Graph Long Short-Term Memory (Graph
LSTM) network, a generalization of LSTM from sequential data or
multidimensional data to general graph-structured data. Instead of
evenly dividing an image to pixels or patches in existing
multi-dimensional LSTM structures (e.g., row, grid and diagonal
LSTMs), they take each arbitrary-shaped superpixel as a semantically
consistent node, and adaptively construct an undirected graph for the
image, where the spatial relations of the superpixels are naturally
used as edges. Figure~\ref{fig:graph_lstm} presents a visual
comparison of the traditional pixel-wise RNN model and graph-LSTM
model. To adapt the Graph LSTM model to semantic segmentation
(Figure~\ref{fig:seg_graph_lstm})
, LSTM layers built on a super-pixel
map are appended on the convolutional layers to enhance visual
features with global structure context. The convolutional features
pass through $1\times1$ convolutional filters to generate the initial
confidence maps for all labels. The node updating sequence for the
subsequent Graph LSTM layers is determined by the confidence-drive
scheme based on the initial confidence maps, and then the Graph LSTM
layers can sequentially update the hidden states of all superpixel
nodes.

Xiang and Fox \cite{DARNN} proposed Data Associated Recurrent Neural
Networks (DA-RNNs), for joint 3D scene mapping and semantic labeling.
DA-RNNs use a new recurrent neural network architecture
for semantic labeling on RGB-D videos. The
output of the network is integrated with mapping techniques such as
Kinect-Fusion in order to inject semantic information into the reconstructed 3D scene.

\begin{figure}[h]
\centering
\includegraphics[width=0.8\linewidth]{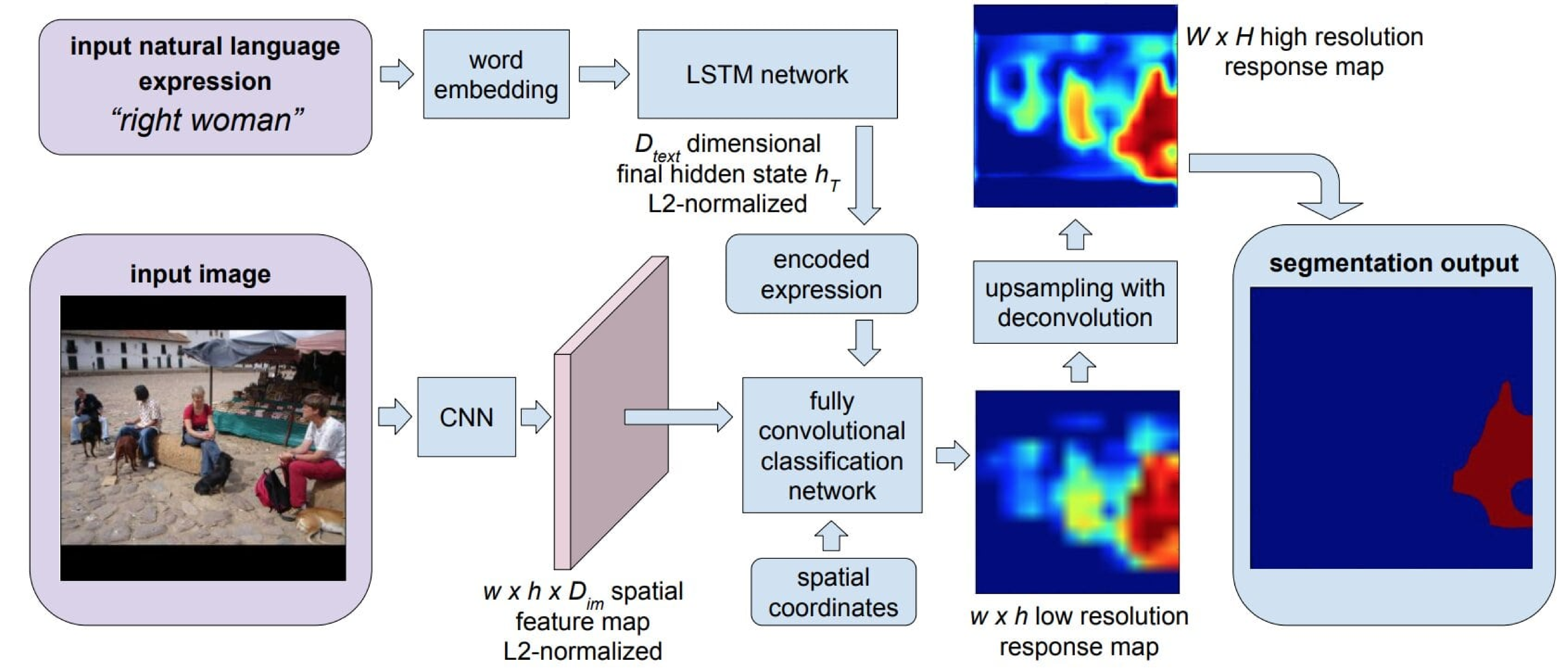}
\caption{The CNN+LSTM architecture for segmentation from natural
language expressions. From \cite{seg_lstm_cnn}.}
\label{fig:lstm_cnn_blk}
\end{figure}

\begin{figure}[h]
\centering
\includegraphics[width=0.8\linewidth]{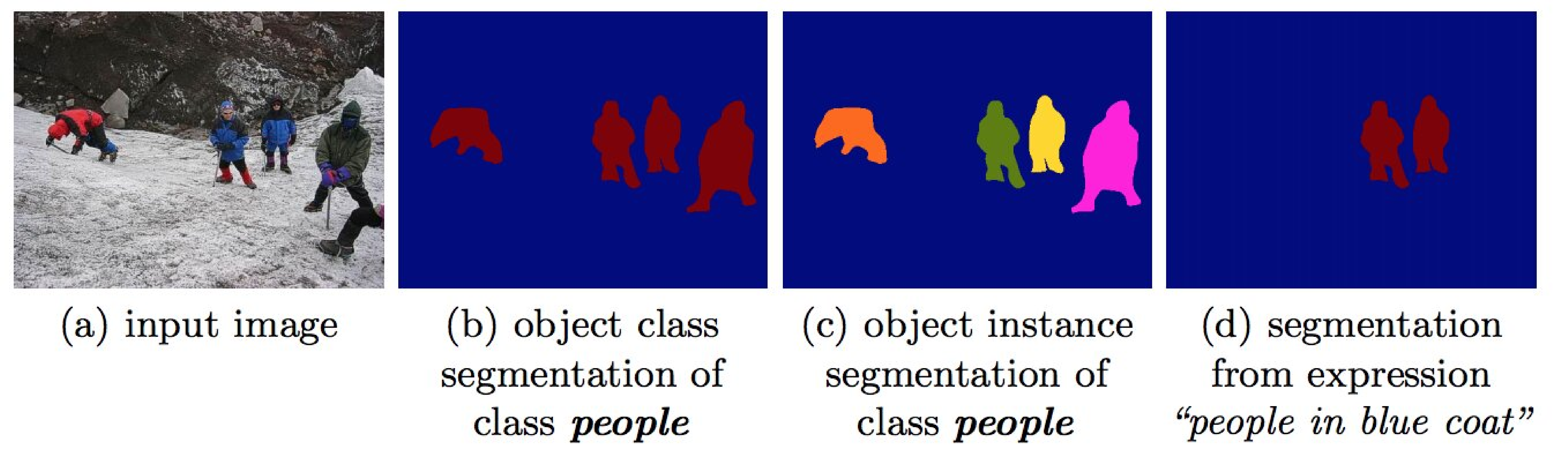}
\caption{Segmentation masks generated for the query ``people in blue
coat''. From \cite{seg_lstm_cnn}.}
\label{fig:seg_query1}
\end{figure}

Hu \textit{et al.} \cite{seg_lstm_cnn} developed a semantic
segmentation algorithm based on natural language expression, using a
combination of CNN to encode the image and LSTM to encode its natural
language description. This is different from traditional semantic
segmentation over a predefined set of semantic classes, as, e.g., the
phrase ``two men sitting on the right bench'' requires segmenting only
the two people on the right bench and no one standing or sitting on
another bench. To produce pixel-wise segmentation for language
expression, they propose an end-to-end trainable recurrent and
convolutional model that jointly learns to process visual and
linguistic information (Figure~\ref{fig:lstm_cnn_blk}). In the
considered model, a recurrent LSTM network is used to encode the
referential expression into a vector representation, and an FCN is
used to extract a spatial feature map from the image and output a
spatial response map for the target object. 
An example segmentation result of this model (for the query ``people in blue coat'') is shown in Figure~\ref{fig:seg_query1}.

One limitation of RNN based models is that, due to the sequential nature these models, they will be slower than their CNN counterpart, since this sequential calculation cannot be parallelized easily.

\subsection{Attention-Based Models}

Attention mechanisms have been persistently explored in computer
vision over the years, and it is therefore not surprising to find
publications that apply such mechanisms to semantic segmentation.

\begin{figure}[h]
\centering
\includegraphics[width=0.7\linewidth]{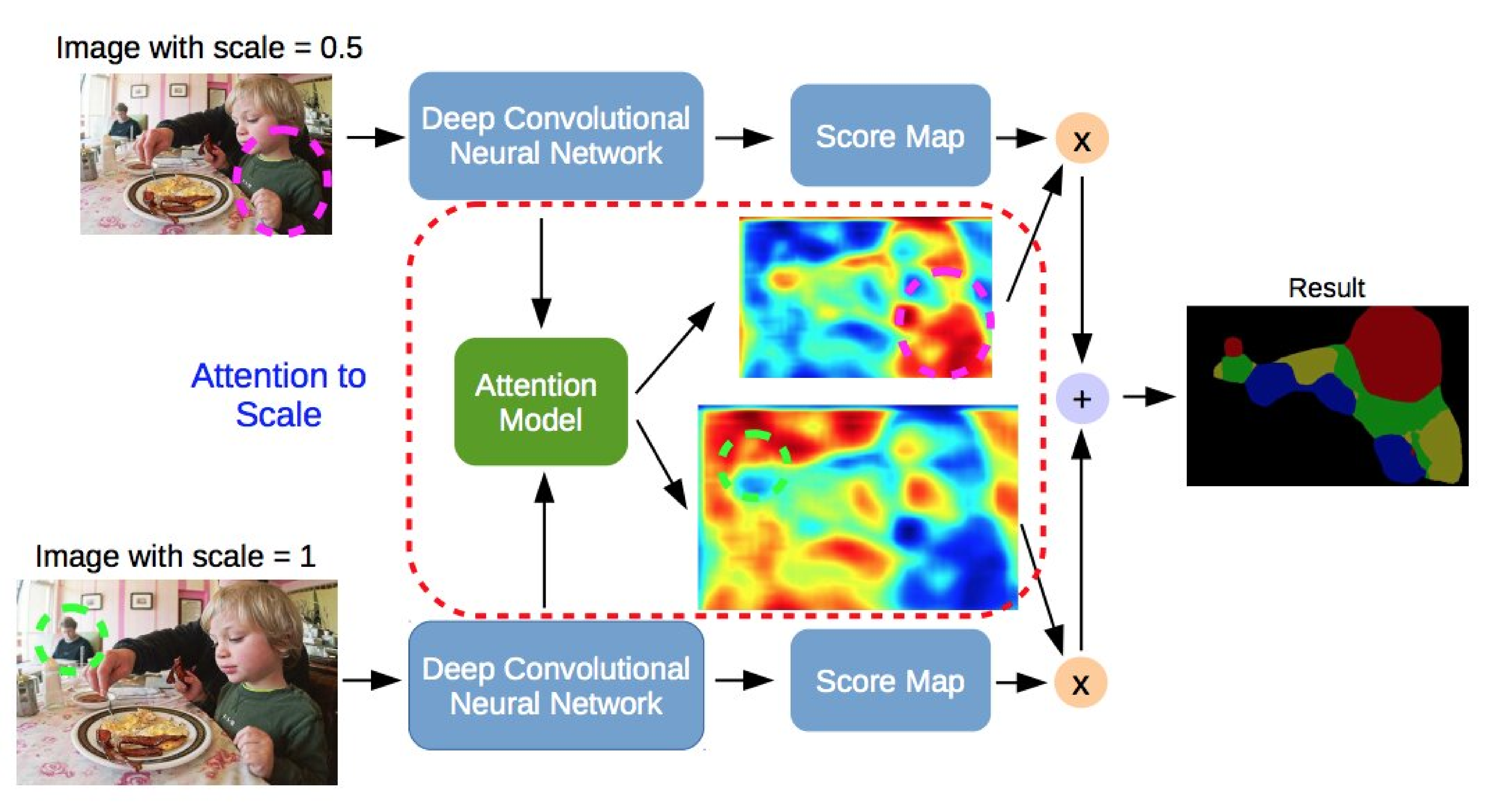}
\caption{Attention-based semantic segmentation model. The attention
model learns to assign different weights to objects of different
scales; e.g., the model assigns large weights on the small person
(green dashed circle) for features from scale 1.0, and large weights
on the large child (magenta dashed circle) for features from scale
0.5. From \cite{seg_att1}.}
\label{fig:seg_att1}
\end{figure}

Chen \textit{et al.} \cite{seg_att1} proposed an attention mechanism
that learns to softly weight multi-scale features at each pixel
location. They adapt a powerful semantic segmentation model and
jointly train it with multi-scale images and the attention model
(Figure~\ref{fig:seg_att1}). The attention mechanism outperforms
average and max pooling, and it enables the model to assess the
importance of features at different positions and scales.

\begin{figure}[h]
\centering
\includegraphics[width=0.7\linewidth]{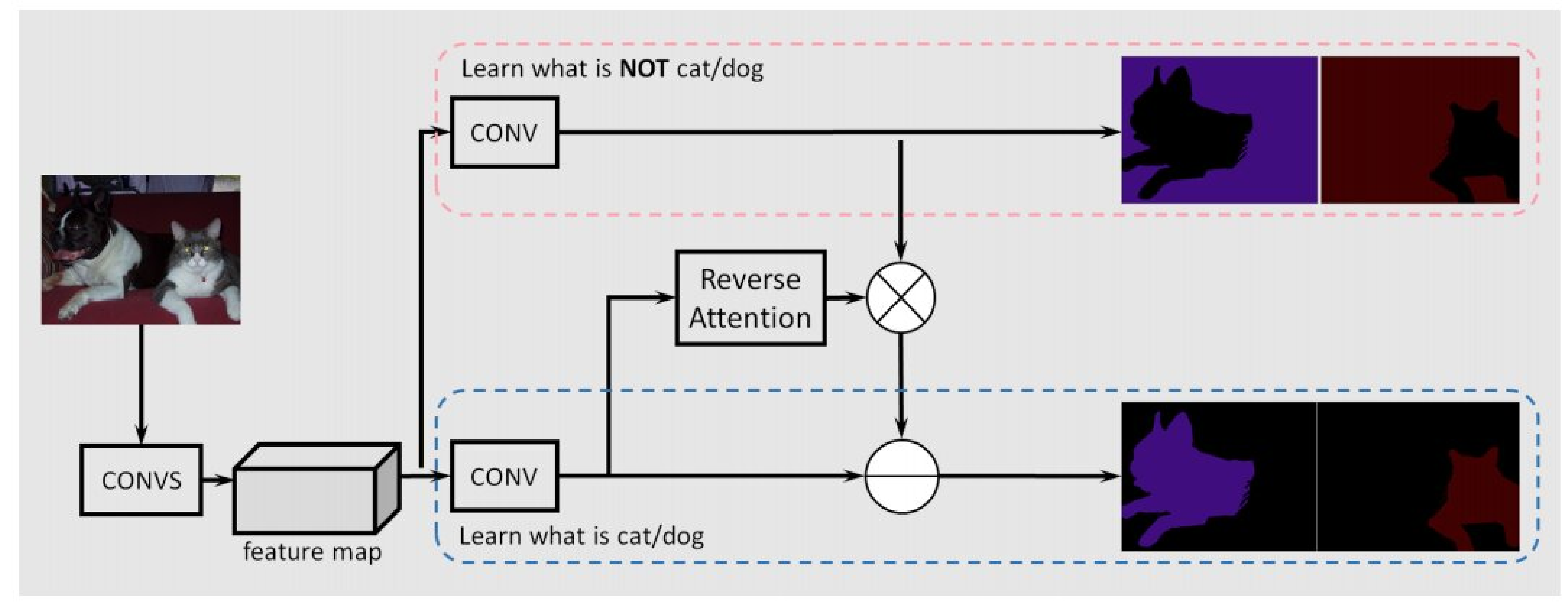}
\caption{The reverse attention network for segmentation. From
\cite{seg_att2}.}
\label{fig:seg_att2}
\end{figure}

In contrast to other works in which convolutional classifiers are
trained to learn the representative semantic features of labeled
objects, Huang \textit{et al.} \cite{seg_att2} proposed a semantic
segmentation approach using reverse attention mechanisms. Their
Reverse Attention Network (RAN) architecture
(Figure~\ref{fig:seg_att2}) trains the model to capture the opposite
concept (i.e., features that are not associated with a target class)
as well. The RAN is a three-branch network that performs the direct,
and reverse-attention learning processes simultaneously.

Li \textit{et al.} \cite{seg_att3} developed a Pyramid Attention
Network for semantic segmentation. This model exploits the impact of
global contextual information in semantic segmentation. They combined
attention mechanisms and spatial pyramids to extract precise dense
features for pixel labeling, instead of complicated dilated
convolutions and artificially designed decoder networks.

More recently, Fu \textit{et al.} \cite{seg_DAN} proposed a dual
attention network for scene segmentation, which can capture rich
contextual dependencies based on the self-attention mechanism.
Specifically, they append two types of attention modules on top of a
dilated FCN which models the semantic inter-dependencies in spatial
and channel dimensions, respectively. The position attention module
selectively aggregates the feature at each position by a weighted sum
of the features at all positions.

Various other works explore attention mechanisms for semantic
segmentation, such as OCNet \cite{OCNet} which proposed an object context pooling inspired by self-attention mechanism, 
Expectation-Maximization Attention (EMANet) \cite{EMAnet}, 
Criss-Cross Attention Network (CCNet) \cite{CcNet},
end-to-end instance segmentation with recurrent attention \cite{seg_rec_att}, a point-wise
spatial attention network for scene parsing \cite{seg_att5}, and a
discriminative feature network (DFN) \cite{seg_att6}, which comprises
two sub-networks: a Smooth Network (that contains a Channel Attention
Block and global average pooling to select the more discriminative
features) and a Border Network (to make the bilateral features of the
boundary distinguishable).

\subsection{Generative Models and Adversarial Training}

Since their introduction, GANs have been applied to a wide range tasks
in computer vision, and have been adopted for image segmentation too.

\begin{figure}
\centering
\includegraphics[width=0.7\linewidth]{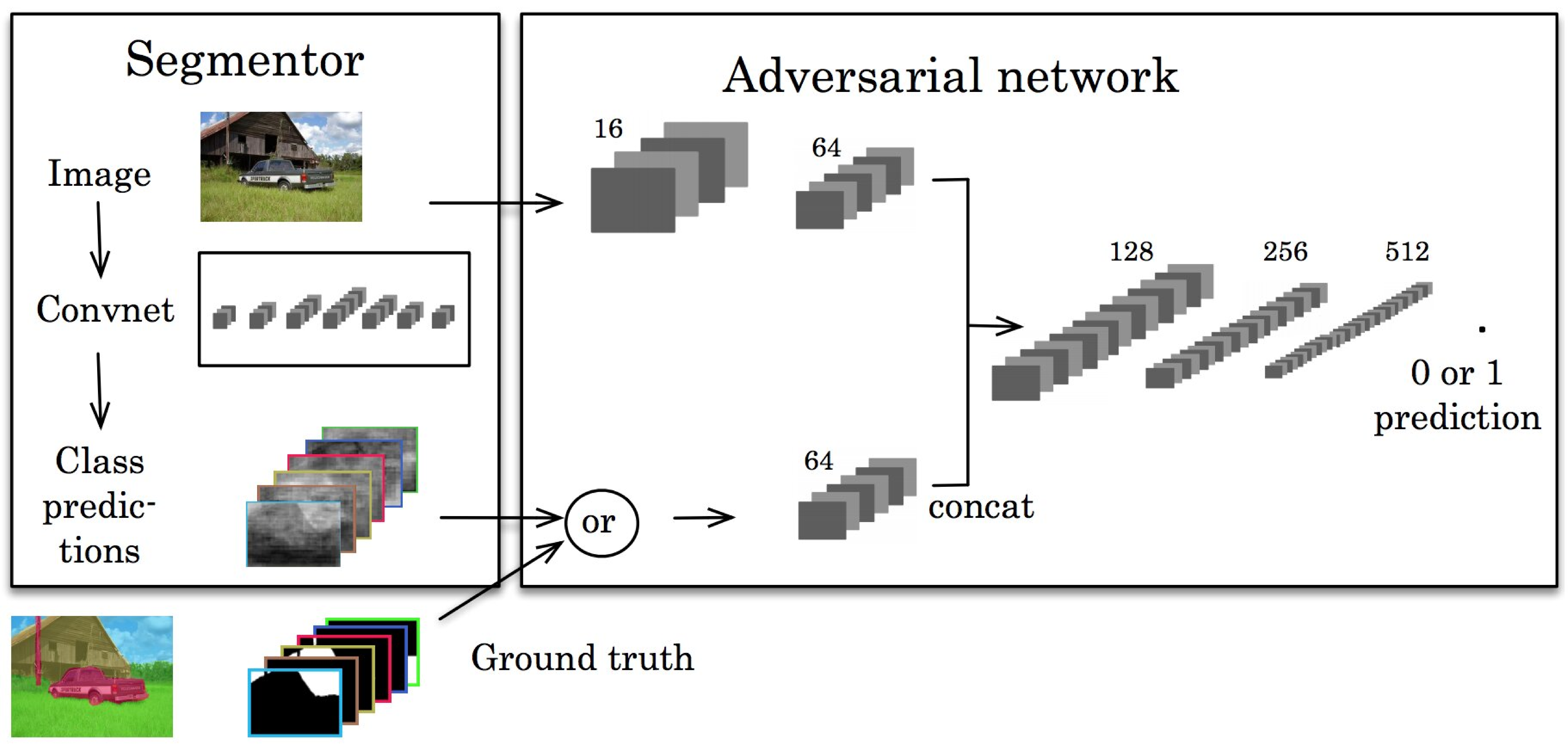}
\caption{The GAN for semantic segmentation. From \cite{seg_gan1}.}
\label{fig:seg_gan1}
\end{figure}

Luc \textit{et al.} \cite{seg_gan1} proposed an adversarial training
approach for semantic segmentation. 
They trained a convolutional
semantic segmentation network (Figure~\ref{fig:seg_gan1}), along with
an adversarial network that discriminates ground-truth segmentation
maps from those generated by the segmentation network. They showed
that the adversarial training approach leads to improved accuracy on
the Stanford Background and PASCAL VOC 2012 datasets.

Souly \textit{et al.} \cite{seg_gan2} proposed semi-weakly supervised
semantic segmentation using GANs. It consists of a generator network
providing extra training examples to a multi-class classifier, acting
as discriminator in the GAN framework, that assigns sample a label y
from the $K$ possible classes or marks it as a fake sample (extra
class).

In another work, Hung \textit{et al.} \cite{seg_gan3} developed a framework for semi-supervised semantic segmentation using an adversarial network.
They designed an FCN discriminator to differentiate the predicted
probability maps from the ground truth segmentation distribution,
considering the spatial resolution. The considered loss function of
this model contains three terms: cross-entropy loss on the
segmentation ground truth, adversarial loss of the discriminator
network, and semi-supervised loss based on the confidence map; i.e.,
the output of the discriminator.

Xue \textit{et al.} \cite{seg_gan4} proposed an adversarial network
with multi-scale L1 Loss for medical image segmentation. They used an
FCN as the segmentor to generate segmentation label maps, and proposed
a novel adversarial critic network with a multi-scale L1 loss function
to force the critic and segmentor to learn both global and local
features that capture long and short range spatial relationships
between pixels.

Various other publications report on segmentation models based on
adversarial training, such as Cell Image Segmentation Using GANs \cite{seg_gan5}, and segmentation and generation of the invisible parts of objects \cite{seg_gan8}.


\subsection{CNN Models With Active Contour Models}

The exploration of synergies between FCNs and Active Contour Models
(ACMs) \cite{Snakes} 
has recently attracted research interest.
One approach is to formulate new loss functions that are inspired by ACM principles. For example, inspired by the global energy formulation of \cite{chan2001active}, Chen \textit{et al.} \cite{chen2019learning} proposed a supervised loss layer that incorporated area and size information of the predicted masks during training of an FCN and tackled the problem of ventricle segmentation in cardiac MRI.


A different approach initially sought to utilize the ACM merely as a post-processor of the output of an FCN and several efforts
attempted modest co-learning by pre-training the FCN. 
One example of an ACM post-processor for the task of semantic
segmentation of natural images is the work by Le \textit{et al.}
\cite{le2018reformulating} in which level-set ACMs are implemented as
RNNs. 
Deep Active Contours by Rupprecht \textit{et al.} \cite{deepacm}, is another example.
For medical image segmentation, Hatamizadeh \textit{et al.} \cite{hatamizadeh2019deep} 
proposed an integrated Deep
Active Lesion Segmentation (DALS) model that trains the FCN backbone
to predict the parameter functions of a novel, locally-parameterized
level-set energy functional. 
In another relevant effort, Marcos
\textit{et al.} \cite{marcos2018learning} proposed Deep Structured
Active Contours (DSAC), which combines ACMs and pre-trained FCNs in a
structured prediction framework for building instance segmentation
(albeit with manual initialization) in aerial images. 
For the same application, Cheng \textit{et al.} \cite{cheng2019darnet} proposed the
Deep Active Ray Network (DarNet), which is similar to DSAC, but with a
different explicit ACM formulation based on polar coordinates to
prevent contour self-intersection.
A truly end-to-end backpropagation trainable, fully-integrated FCN-ACM
combination was recently introduced by Hatamizadeh \textit{et al.}
\cite{hatamizadeh2019end}, dubbed Deep Convolutional Active Contours
(DCAC). 


\subsection{Other Models}
In addition to the above models, there are several other popular DL
architectures for segmentation, such as the following: Context
Encoding Network (EncNet) that uses a basic feature extractor and feeds the feature maps into a Context Encoding Module \cite{EncNet}.
RefineNet \cite{Refinenet}, which is a multi-path refinement network that explicitly exploits all the information available along the down-sampling process to enable high-resolution prediction using
long-range residual connections. 
Seednet \cite{Seednet}, which introduced an automatic seed generation
technique with deep reinforcement learning that learns to solve the
interactive segmentation problem.
"Object-Contextual Representations" (OCR) \cite{hrrocr}, which learns object regions under the supervision of the ground-truth, and computes the object region representation, and the relation between each pixel
and each object region, and augment the representation pixels with the object-contextual representation. 
Yet additional models include BoxSup  \cite{Boxsup}, Graph convolutional networks \cite{GCN}, Wide ResNet \cite{wideresnet}, Exfuse (enhancing low-level and high-level features fusion) \cite{Exfuse}, Feedforward-Net \cite{seg_feedforward}, saliency-aware models for geodesic video segmentation \cite{sal_Seg},
dual image segmentation (DIS) \cite{DIS}, FoveaNet (Perspective-aware scene parsing) \cite{Foveanet}, Ladder DenseNet \cite{ladder},  Bilateral segmentation network (BiSeNet) \cite{BiSeNet}, Semantic Prediction Guidance for Scene Parsing (SPGNet) \cite{SPGNet}, Gated shape CNNs \cite{GSCNN}, Adaptive context network (AC-Net) \cite{ACnet}, Dynamic-structured semantic propagation network (DSSPN) \cite{DSSPN}, symbolic graph reasoning (SGR) \cite{SGR}, CascadeNet \cite{ADE20k}, Scale-adaptive convolutions (SAC) \cite{SAC}, Unified perceptual parsing (UperNet) \cite{uper}, segmentation by re-training and self-training \cite{zoph2020rethinking}, densely connected neural architecture search \cite{zhang2020dcnas}, hierarchical multi-scale attention \cite{tao2020hierarchical}.

Panoptic segmentation \cite{kirillov2019panoptic} is also another interesting segmentation problem with rising popularity, and there are already several interesting works on this direction, including Panoptic Feature Pyramid Network \cite{pfpn}, attention-guided network for Panoptic segmentation \cite{seg_att4}, Seamless Scene Segmentation \cite{porzi2019seamless}, panoptic deeplab \cite{cheng2019panoptic}, unified panoptic segmentation network \cite{xiong2019upsnet}, efficient panoptic segmentation \cite{mohan2020efficientps}.

\begin{figure*}
\centering
\includegraphics[width=0.95\linewidth]{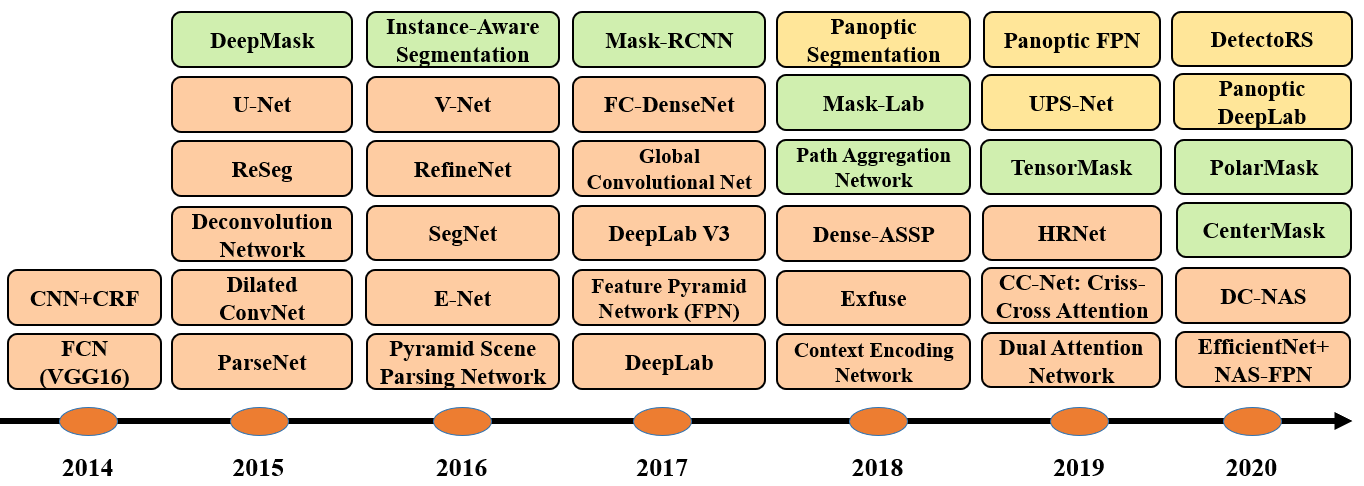}
\caption{The timeline of DL-based segmentation algorithms for 2D images, from 2014 to 2020. Orange, green, andn yellow blocks refer to semantic, instance, and panoptic segmentation algorithms respectively.}
\label{seg_timeline}
\end{figure*}

Figure \ref{seg_timeline} illustrates the timeline of popular DL-based works for semantic segmentation, as well as instance segmentation since 2014.
Given the large number of works developed in the last few years, we only show some of the most representative ones.

\section{Image Segmentation Datasets}
\label{sec:datasets}

In this section we provide a summary of some of the most widely used image segmentation
datasets. We group these datasets into 3 categories---2D images, 2.5D
RGB-D (color+depth) images, and 3D images---and provide details about
the characteristics of each dataset. The listed datasets have pixel-wise labels, which can be used for evaluating model performance.

It is worth mentioning that some of these works, use \textbf{data augmentation} to increase the number of labeled samples, specially the ones which deal with small datasets (such as in medical domain).
Data augmentation serves to increase the number of training samples by
applying a set of transformation (either in the data space, or feature
space, or sometimes both) to the images (i.e., both the input image
and the segmentation map). Some typical transformations include
translation, reflection, rotation, warping, scaling, color space
shifting, cropping, and projections onto principal components. Data
augmentation has proven to improve the performance of the models,
especially when learning from limited datasets, such as those in
medical image analysis. 
It can also be beneficial in yielding faster
convergence, decreasing the chance of over-fitting, and enhancing
generalization. 
For some small datasets, data augmentation has been shown to boost model performance more than 20\%.

\subsection{2D Datasets}

\begin{figure}[h]
\centering
\includegraphics[width=0.7\linewidth]{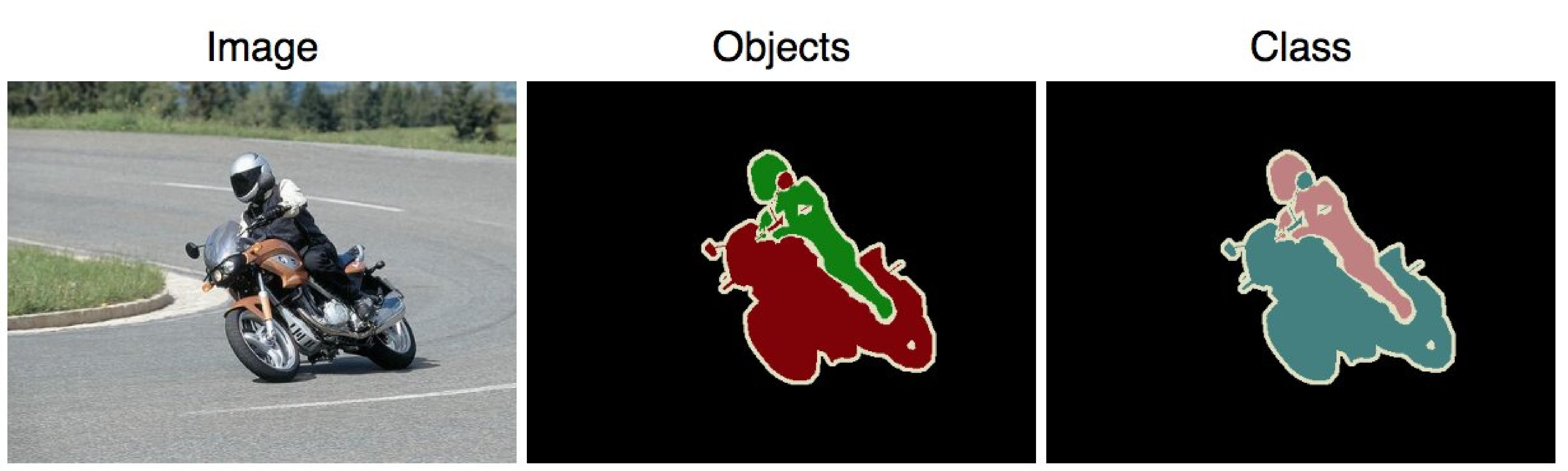}
\caption{An example image from the PASCAL VOC dataset. From
\cite{pascal_site}}
\label{fig:pascal_voc2010}
\end{figure}

The majority of image segmentation research has focused on 2D images;
therefore, many 2D image segmentation datasets are available. The
following are some of the most popular:

\textbf{PASCAL Visual Object Classes (VOC)} \cite{pascal_voc2010} is
one of most popular datasets in computer vision, with annotated images
available for 5 tasks---classification, segmentation, detection,
action recognition, and person layout. Nearly all popular segmentation
algorithms reported in the literature have been evaluated on this
dataset. For the segmentation task, there are 21 classes of object
labels---vehicles, household, animals, aeroplane, bicycle, boat, bus,
car, motorbike, train, bottle, chair, dining table, potted plant,
sofa, TV/monitor, bird, cat, cow, dog, horse, sheep, and person (pixel
are labeled as background if they do not belong to any of these
classes). This dataset is divided into two sets, training and
validation, with 1,464 and 1,449 images, respectively. There is a
private test set for the actual challenge.
Figure~\ref{fig:pascal_voc2010} shows an example image and its
pixel-wise label.

\textbf{PASCAL Context} \cite{pascal_context} is an extension of the
PASCAL VOC 2010 detection challenge, and it contains pixel-wise labels
for all training images. It contains more than 400 classes (including
the original 20 classes plus backgrounds from PASCAL VOC
segmentation), divided into three categories (objects, stuff, and
hybrids). Many of the object categories of this dataset are too sparse
and; therefore, a subset of 59 frequent classes are usually selected
for use. 

\begin{figure}[h]
\centering
\includegraphics[width=0.5\linewidth]{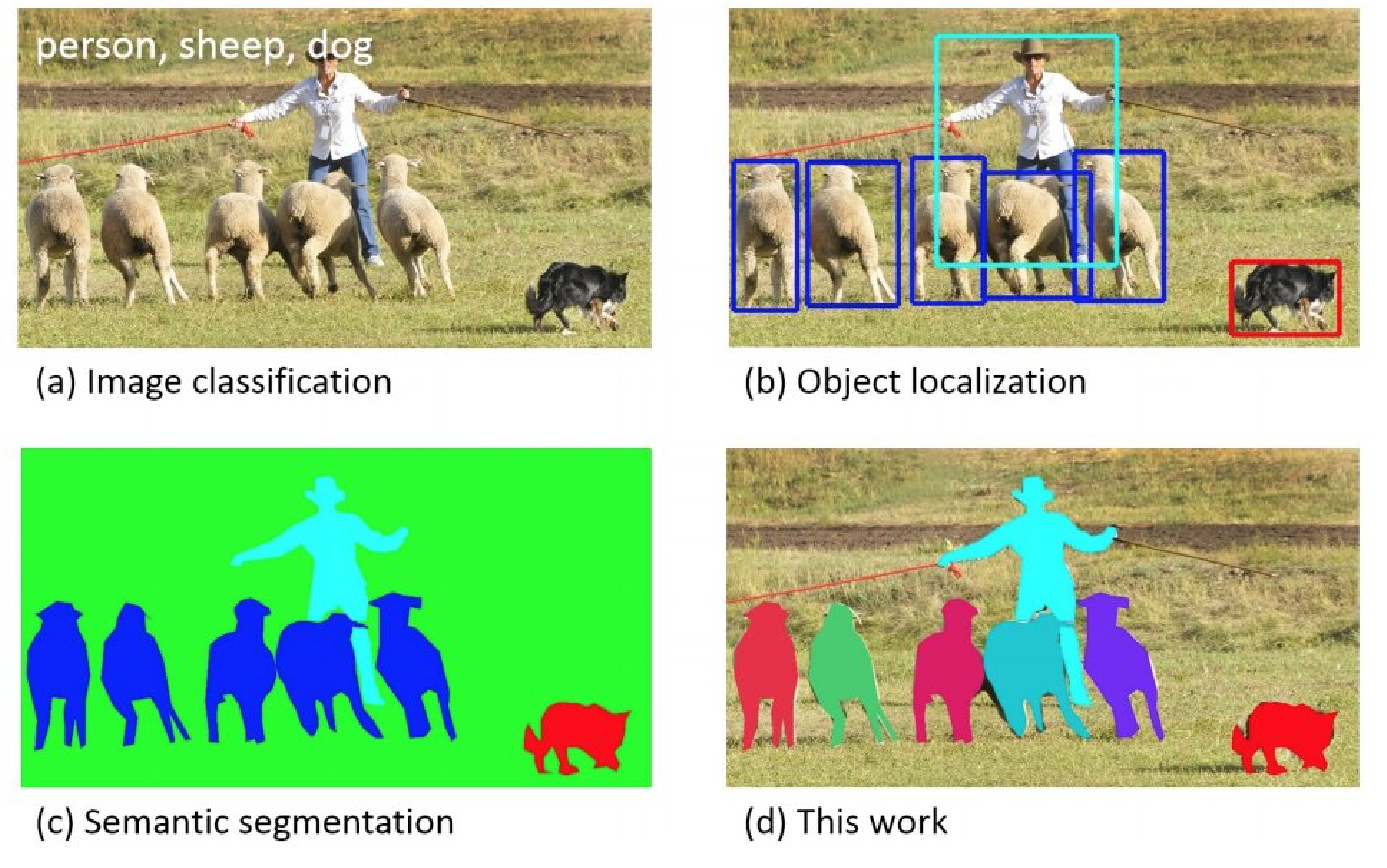}
\caption{A sample image and its segmentation map in COCO, and its
comparison with previous datasets. From \cite{mscoco}.}
\label{fig:mscoco}
\end{figure}

\textbf{Microsoft Common Objects in Context (MS COCO)} \cite{mscoco}
is another large-scale object detection, segmentation, and captioning
dataset. COCO includes images of complex everyday scenes, containing
common objects in their natural contexts. This dataset contains photos
of 91 objects types, with a total of 2.5 million labeled instances in
328k images. 
Figure~\ref{fig:mscoco} shows the difference between MS-COCO labels and the previous datasets for a given sample image. 
The detection challenge includes more than 80 classes, providing more than
82k images for training, 40.5k images for validation, and more than 80k images for its test set.

\begin{figure}[h]
\centering
\includegraphics[width=0.99\linewidth]{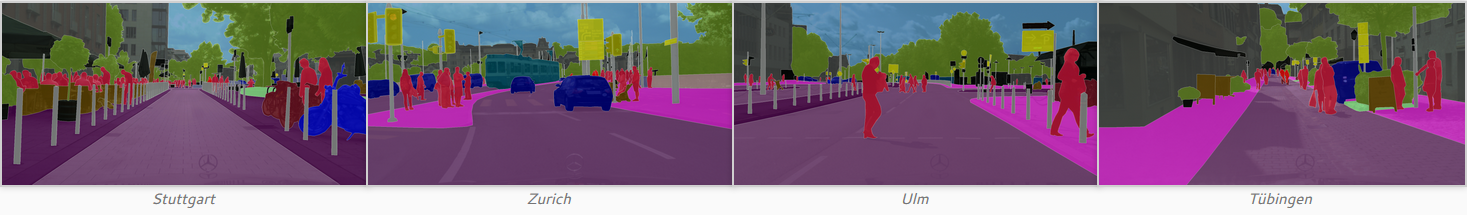}
\caption{Three sample images with their corresponding segmentation
maps from the Cityscapes dataset. From \cite{Cityscapes}.}
\label{fig:Cityscapes}
\end{figure}

\textbf{Cityscapes} \cite{Cityscapes} is a large-scale database with a
focus on semantic understanding of urban street scenes. 
It contains a diverse set of stereo video sequences recorded in street scenes from 50 cities, with high quality pixel-level annotation of
5k frames, in addition to a set of 20k weakly annotated frames. It includes semantic and dense pixel annotations of 30
classes, grouped into 8 categories---flat surfaces, humans, vehicles,
constructions, objects, nature, sky, and void.
Figure~\ref{fig:Cityscapes} shows four sample segmentation maps from this dataset.

\textbf{ADE20K\,/\,MIT Scene Parsing (SceneParse150)} offers a
standard training and evaluation platform for scene parsing
algorithms. The data for this benchmark comes from the ADE20K dataset
\cite{ADE20k}, which contains more than 20K scene-centric images
exhaustively annotated with objects and object parts. The benchmark is
divided into 20K images for training, 2K images for validation, and
another batch of images for testing. There are 150 semantic categories in this dataset.

\textbf{SiftFlow} \cite{SiftFlow} includes 2,688 annotated images from
a subset of the LabelMe database. The $256\times256$ pixel images are
based on 8 different outdoor scenes, among them streets, mountains,
fields, beaches, and buildings. All images belong to one of 33
semantic classes.

\textbf{Stanford background} \cite{stan_back} contains outdoor images
of scenes from existing datasets, such as LabelMe, MSRC, and PASCAL
VOC. It contains 715 images with at least one foreground object. The
dataset is pixel-wise annotated, and can be used for semantic scene
understanding. Semantic and geometric labels for this dataset were
obtained using Amazon's Mechanical Turk (AMT).

\textbf{Berkeley Segmentation Dataset (BSD)} \cite{BSD} contains
12,000 hand-labeled segmentations of 1,000 Corel dataset images from
30 human subjects. It aims to provide an empirical basis for research
on image segmentation and boundary detection. 
Half of the segmentations were obtained from presenting the subject a color image
and the other half from presenting a grayscale image. 

\textbf{Youtube-Objects} \cite{youtube_video} contains videos
collected from YouTube, which include objects from ten PASCAL VOC
classes (aeroplane, bird, boat, car, cat, cow, dog, horse, motorbike,
and train). The original dataset did not contain pixel-wise
annotations (as it was originally developed for object detection, with
weak annotations). However, Jain \textit{et al.} \cite{jain_youtube}
manually annotated a subset of 126 sequences, and then extracted a
subset of frames to further generate semantic labels. In total, there
are about 10,167 annotated 480x360 pixel frames available in this
dataset.


\textbf{KITTI} \cite{kitti} is one of the most popular datasets for
mobile robotics and autonomous driving. It contains hours of videos of
traffic scenarios, recorded with a variety of sensor modalities
(including high-resolution RGB, grayscale stereo cameras, and a 3D
laser scanners). The original dataset does not contain ground truth
for semantic segmentation, but researchers have manually annotated
parts of the dataset for research purposes. For example, Alvarez
\textit{et al.} \cite{Alvarez} generated ground truth for 323 images
from the road detection challenge with 3 classes, road, vertical,
and sky.

\textbf{Other Datasets} are available for image segmentation purposes too,
such as \textbf{Semantic Boundaries Dataset (SBD)} \cite{sbd},
\textbf{PASCAL Part} \cite{pascal_part}, \textbf{SYNTHIA}
\cite{synthia}, and \textbf{Adobe’s Portrait Segmentation}
\cite{adobe}.

\subsection{2.5D Datasets}

\begin{figure}[h]
\centering
   \includegraphics[width=0.8\linewidth]{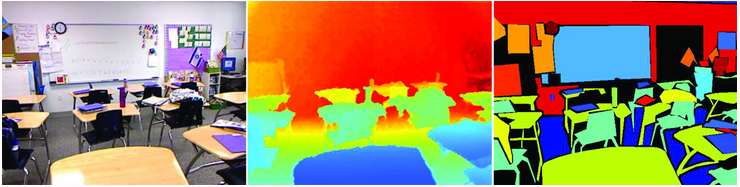}
   \caption{A sample from the NYU V2 dataset. From left: the RGB
   image, pre-processed depth, and set of labels. From \cite{nyuv2}.}
\label{fig:nyuv2}
\end{figure}

With the availability of affordable range scanners, RGB-D images have
became popular in both research and industrial applications. The
following RGB-D datasets are some of the most popular:

\textbf{NYU-D V2} \cite{nyuv2} consists of video sequences from a
variety of indoor scenes, recorded by the RGB and depth cameras of the
Microsoft Kinect. It includes 1,449 densely labeled pairs of aligned
RGB and depth images from more than 450 scenes taken from 3 cities.
Each object is labeled with a class and an instance number (e.g.,
cup1, cup2, cup3, etc.). It also contains 407,024 unlabeled frames.
This dataset is relatively small compared to other existing datasets.
Figure~\ref{fig:nyuv2} shows a sample image and its segmentation map.

\textbf{SUN-3D} \cite{sun3d} is a large-scale RGB-D video dataset that
contains 415 sequences captured for 254 different spaces in 41
different buildings; 8 sequences are annotated and more will be
annotated in the future. Each annotated frame comes with the semantic
segmentation of the objects in the scene, as well as information about
the camera pose.

\textbf{SUN RGB-D} \cite{sunrgbd} provides an RGB-D benchmark for the
goal of advancing the state-of-the-art in all major scene
understanding tasks. It is captured by four different sensors and
contains 10,000 RGB-D images at a scale similar to PASCAL VOC. The
whole dataset is densely annotated and includes 146,617 2D polygons
and 58,657 3D bounding boxes with accurate object orientations, as
well as the 3D room category and layout for scenes. 

\textbf{UW RGB-D Object Dataset} \cite{uw_rgbd} contains 300 common
household objects recorded using a Kinect style 3D camera. The objects
are organized into 51 categories, arranged using WordNet
hypernym-hyponym relationships (similar to ImageNet). This dataset was
recorded using a Kinect style 3D camera that records synchronized and
aligned $640\times480$ pixel RGB and depth images at 30\,Hz.
This dataset also includes 8 annotated video sequences of natural
scenes, containing objects from the dataset (the UW RGB-D Scenes
Dataset).

\textbf{ScanNet} \cite{scannet} is an RGB-D video dataset containing
2.5 million views in more than 1,500 scans, annotated with 3D camera
poses, surface reconstructions, and instance-level semantic
segmentations. To collect these data, an easy-to-use and scalable
RGB-D capture system was designed that includes automated surface
reconstruction, and the semantic annotation was crowd-sourced. Using
this data helped achieve state-of-the-art performance on several 3D
scene understanding tasks, including 3D object classification,
semantic voxel labeling, and CAD model retrieval.

\subsection{3D Datasets}

3D image datasets are popular in  robotic, medical image analysis, 3D scene analysis, and construction applications.
Three dimensional images are usually provided via meshes or other volumetric representations, such as point clouds.
Here, we mention some of the popular 3D datasets.

\textbf{Stanford 2D-3D:}
This dataset provides a variety of mutually registered modalities from 2D, 2.5D and 3D domains, with instance-level semantic and geometric annotations \cite{stan3d}, and is collected in 6 indoor areas. It contains over 70,000 RGB images, along with the corresponding depths, surface normals, semantic annotations, global XYZ images as well as camera information. 

\textbf{ShapeNet Core:}
ShapeNetCore is a subset of the full ShapeNet dataset \cite{shapenet} with single clean 3D models and manually verified category and alignment annotations \cite{shapenetcore}. 
It covers 55 common object categories with about 51,300 unique 3D models. 

\textbf{Sydney Urban Objects Dataset:}
This dataset contains a variety of common urban road objects, collected in the central business district of Sydney, Australia. 
There are 631 individual scans of objects across classes of vehicles, pedestrians, signs and trees \cite{sydney3d}.

\section{Performance Review}
\label{sec:performance}

In this section, we first provide a summary of some of the popular metrics used in evaluating the performance of segmentation models, and then we provide the quantitative performance of the promising DL-based segmentation models on popular datasets.

\subsection{Metrics For Segmentation Models}
\label{sec:metrics}

Ideally, a model should be evaluated in
multiple respects, such as quantitative accuracy, speed (inference
time), and storage requirements (memory footprint). 
However, most of the research works so far, focus on the metrics for evaluating the model accuracy.
Below we summarize the most popular
metrics for assessing the accuracy of segmentation algorithms.
Although quantitative metrics are used to compare different models on
benchmarks, the visual quality of model outputs is also important in
deciding which model is best (as human is the final consumer of many of the models developed for computer vision applications).

\textbf{Pixel accuracy} simply finds the ratio of pixels properly
classified, divided by the total number of pixels. For $K+1$ classes
($K$ foreground classes and the background) pixel accuracy is defined as Eq \ref{eq_PA}:
\begin{equation}
\text{PA}= \frac{\sum_{i=0}^K p_{ii}} {\sum_{i=0}^K \sum_{j=0}^K p_{ij}},
\label{eq_PA}
\end{equation}
where $p_{ij}$ is the number of pixels of class $i$ predicted as
belonging to class $j$.

\textbf{Mean Pixel Accuracy (MPA)} is the extended version of PA, in
which the ratio of correct pixels is computed in a per-class manner
and then averaged over the total number of classes, as in Eq \ref{eq_MPA}:
\begin{equation}
\text{MPA}= \frac{1}{K+1} \sum_{i=0}^K \frac{p_{ii}}{\sum_{j=0}^K p_{ij}}.
\label{eq_MPA}
\end{equation}

\textbf{Intersection over Union (IoU)} or the \textbf{Jaccard Index}
is one of the most commonly used metrics in semantic segmentation. It
is defined as the area of intersection between the predicted
segmentation map and the ground truth, divided by the area of union
between the predicted segmentation map and the ground truth:
\begin{equation}
\text{IoU}= J(A,B) =  \frac{|A \cap B|}{|A \cup B|}, 
\end{equation}
where $A$ and $B$ denote the ground truth and the predicted
segmentation maps, respectively. 
It ranges between 0 and 1.

\textbf{Mean-IoU} is another popular metric, which is defined as the
average IoU over all classes. It is widely used in reporting the
performance of modern segmentation algorithms.

\textbf{Precision\,/\,Recall\,/\,F1 score} are popular metrics for
reporting the accuracy of many of the classical image segmentation
models. Precision and recall can be defined for each class, as well as
at the aggregate level, as follows:
\begin{equation}
\begin{aligned}
\text{Precision}&=  \frac{\text{TP}}{\text{TP}+\text{FP}},
\ \ \text{Recall}&=  \frac{\text{TP}}{\text{TP}+\text{FN}},
\end{aligned}
\label{prec_rec}
\end{equation}
where TP refers to the true positive fraction, FP refers to the false
positive fraction, and FN refers to the false negative fraction.
Usually we are interested into a combined version of precision and
recall rates. A popular such a metric is called the F1 score, which is
defined as the harmonic mean of precision and recall:
\begin{equation}
\text{F1-score}=  \frac{2 \ \text{Prec} \ \text{Rec}}{ \text{Prec}+\text{Rec} }.
\label{f1_score}
\end{equation}

\textbf{Dice coefficient} is another popular metric for image
segmentation (and is more commonly used in medical image analysis), which can be defined as twice the overlap area of
predicted and ground-truth maps, divided by the total number of pixels
in both images. 
The Dice coefficient is very similar to the IoU:
\begin{equation}
\text{Dice}= \frac{ 2| A \cap B |}{|A|+ |B|}.
\label{eq_dice}
\end{equation}
When applied to boolean data (e.g., binary segmentation maps), and
referring to the foreground as a positive class, the Dice coefficient
is essentially identical to the F1 score, defined as Eq \ref{eq_dice2}:
\begin{equation}
\text{Dice}= \frac{ 2 \text{TP}}{ 2\text{TP}+ \text{FP}+\text{FN} }=
\text{F1}.
\label{eq_dice2}
\end{equation}

\subsection{Quantitative Performance of DL-Based Models}
\label{sec:quant_result}

In this section we tabulate the performance of several of the
previously discussed algorithms on popular segmentation benchmarks.
It is worth mentioning that although most models report their
performance on standard datasets and use standard metrics, some of
them fail to do so, making across-the-board comparisons difficult.
Furthermore, only a small percentage of publications provide
additional information, such as execution time and memory footprint,
in a reproducible way, which is important to industrial applications
of segmentation models (such as drones, self-driving cars, robotics,
etc.) that may run on embedded consumer devices with limited
computational power and storage, making fast, light-weight models
crucial.








\begin{table}
\centering
\caption{Accuracies of segmentation models on the PASCAL VOC test
set. (*~Refers to the model pre-trained on another dataset (such as MS-COCO, ImageNet, or JFT-300M).)}
\label{table_voc}
\begin{tabular}{llr}
\toprule
Method & Backbone  &  mIoU \\
\midrule
FCN \cite{seg_fcn}  & VGG-16  & 62.2 \\ 
CRF-RNN  \cite{CRF-RNN}  & -  & 72.0 \\  
CRF-RNN$^{*}$  \cite{CRF-RNN}  & -  & 74.7 \\ 
BoxSup*  \cite{Boxsup}  & -  & 75.1 \\ 
Piecewise$^{*}$  \cite{seg_dsn2}  & -  & 78.0 \\ 
DPN$^{*}$  \cite{seg_dsn3}  & -  & 77.5 \\ 
DeepLab-CRF \cite{deeplab}  & ResNet-101  & 79.7 \\
GCN$^{*}$ \cite{GCN}  & ResNet-152  & 82.2 \\ 
RefineNet \cite{Refinenet}  & ResNet-152  & 84.2 \\ 
Wide ResNet \cite{wideresnet}  & WideResNet-38  & 84.9 \\ 
PSPNet \cite{pspn}  & ResNet-101  & 85.4 \\ 
DeeplabV3 \cite{deeplabv3}  & ResNet-101  & 85.7 \\ 
PSANet \cite{seg_att5}  & ResNet-101  & 85.7 \\ 
EncNet \cite{EncNet}  & ResNet-101  & 85.9 \\ 
DFN$^{*}$ \cite{seg_att6}  & ResNet-101  & 86.2 \\ 
Exfuse \cite{Exfuse}  & ResNet-101  & 86.2 \\ 
SDN* \cite{SDN}  & DenseNet-161  & 86.6 \\ 
DIS \cite{DIS}  & ResNet-101  & 86.8 \\ 
DM-Net$^{*}$ \cite{dmsf}  & ResNet-101  & 87.06 \\  
APC-Net$^{*}$ \cite{apcnet}  & ResNet-101  & 87.1 \\  
EMANet \cite{EMAnet}  & ResNet-101  & 87.7 \\ 
DeeplabV3+ \cite{deeplabv3plus}  & Xception-71  & 87.8 \\ 
Exfuse \cite{Exfuse}  & ResNeXt-131  & 87.9 \\ 
MSCI \cite{MSCI}  & ResNet-152  & 88.0 \\
EMANet \cite{EMAnet}  & ResNet-152  & 88.2 \\
DeeplabV3+$^{*}$ \cite{deeplabv3plus}  & Xception-71  & 89.0 \\
EfficientNet+NAS-FPN  \cite{zoph2020rethinking}  & -  & 90.5 \\
\bottomrule
\end{tabular}
\end{table}

The following tables summarize the performances of several of the
prominent DL-based segmentation models on different datasets.
Table~\ref{table_voc} focuses on the PASCAL VOC test set. Clearly,
there has been much improvement in the accuracy of the models since
the introduction of the FCN, the first DL-based image segmentation
model. 
Table~\ref{table_cityscape} focuses on the Cityscape test dataset. The latest models feature about 23\% relative gain
over the initial FCN model on this dataset. Table~\ref{table_mscoco}
focuses on the MS COCO stuff test set. This dataset is more
challenging than PASCAL VOC, and Cityescapes, as the highest mIoU is approximately 40\%. Table~\ref{table_ADE20k} focuses on the ADE20k validation set. This dataset is also more challenging than the PASCAL
VOC and Cityescapes datasets. 

Table~\ref{instance_coco2017} provides the performance of prominent instance segmentation algorithms on COCO test-dev 2017 dataset, in terms of average precision, and their speed.
Table~\ref{panoptic_coco2017} provides the performance of prominent panoptic segmentation algorithms on MS-COCO val dataset, in terms of panoptic quality \cite{kirillov2019panoptic}. 
Finally, Table \ref{table_NYUv2}
summarizes the performance of several prominent models for RGB-D
segmentation on the NYUD-v2 and SUN-RGBD datasets.

\begin{table}
\centering
\caption{Accuracies of segmentation models on the Cityescapes dataset.}
\label{table_cityscape}
\begin{tabular}{llr}
\toprule
Method & Backbone  &  mIoU \\
\midrule
FCN-8s \cite{seg_fcn}  & -  & 65.3 \\ 
DPN \cite{seg_dsn3}  & -  & 66.8 \\ 
Dilation10 \cite{multi_cont_agg}  & -  & 67.1 \\ 
DeeplabV2 \cite{deeplab}  & ResNet-101  & 70.4 \\ 
RefineNet \cite{Refinenet}  & ResNet-101  & 73.6 \\ 
FoveaNet \cite{Foveanet}  & ResNet-101  & 74.1 \\
Ladder DenseNet \cite{ladder}  & Ladder DenseNet-169  & 73.7 \\ 
GCN \cite{GCN}  & ResNet-101  & 76.9 \\ 
DUC-HDC \cite{UCS}  & ResNet-101  & 77.6 \\ 
Wide ResNet \cite{wideresnet}  & WideResNet-38  & 78.4 \\ 
PSPNet \cite{pspn}  & ResNet-101  & 85.4 \\ 
BiSeNet \cite{BiSeNet}  & ResNet-101  & 78.9 \\ 
DFN \cite{seg_att6}  & ResNet-101  & 79.3 \\ 
PSANet \cite{seg_att5}  & ResNet-101  & 80.1 \\ 
DenseASPP \cite{Denseaspp}  & DenseNet-161  & 80.6 \\ 
SPGNet \cite{SPGNet} & 2xResNet-50  & 81.1 \\ 
DANet \cite{seg_DAN}  & ResNet-101  & 81.5 \\ 
CCNet \cite{CcNet}  & ResNet-101  & 81.4 \\ 
DeeplabV3 \cite{deeplabv3}  & ResNet-101  & 81.3 \\ 
AC-Net \cite{ACnet}  & ResNet-101  & 82.3 \\
OCR \cite{hrrocr} & ResNet-101  & 82.4 \\
GS-CNN \cite{GSCNN} & WideResNet  & 82.8 \\
HRNetV2+OCR (w/ASPP) \cite{hrrocr} & HRNetV2-W48 & 83.7\\
Hierarchical MSA \cite{tao2020hierarchical} & HRNet-OCR & 85.1\\
\bottomrule
\end{tabular}
\end{table}

\begin{table}
\centering
\caption{Accuracies of segmentation models on the MS COCO stuff
dataset.}
\label{table_mscoco}
\begin{tabular}{llr}
\toprule
Method & Backbone  &  mIoU \\
\midrule
RefineNet \cite{Refinenet}  & ResNet-101  & 33.6 \\  
CCN \cite{CCL}  & Ladder DenseNet-101  & 35.7 \\
DANet \cite{seg_DAN}  & ResNet-50  & 37.9 \\ 
DSSPN \cite{DSSPN}  & ResNet-101  & 37.3 \\  
EMA-Net \cite{EMAnet}  & ResNet-50  & 37.5 \\  
SGR \cite{SGR}  & ResNet-101  & 39.1 \\ 
OCR \cite{hrrocr}  & ResNet-101  & 39.5 \\
DANet \cite{seg_DAN}  & ResNet-101  & 39.7 \\ 
EMA-Net \cite{EMAnet}  & ResNet-50  & 39.9 \\ 
AC-Net \cite{ACnet}  & ResNet-101  & 40.1 \\
OCR \cite{hrrocr}  & HRNetV2-W48  & 40.5 \\
\bottomrule
\end{tabular}
\end{table}

\begin{table}
\centering
\caption{Accuracies of segmentation models on the ADE20k validation
dataset.}
\label{table_ADE20k}
\begin{tabular}{llr}
Method & Backbone  &  mIoU \\
\midrule
FCN \cite{seg_fcn}  &  - & 29.39 \\  
DilatedNet \cite{multi_cont_agg}  & - & 32.31 \\  
CascadeNet \cite{ADE20k}  &  -  & 34.9 \\ RefineNet \cite{Refinenet}  &  ResNet-152  & 40.7 \\ 
PSPNet \cite{pspn}  & ResNet-101  &  43.29 \\  
PSPNet \cite{pspn}  & ResNet-269  &  44.94 \\  
EncNet \cite{EncNet}  & ResNet-101  &  44.64 \\  
SAC \cite{SAC}  & ResNet-101  & 44.3 \\  
PSANet \cite{seg_att5}  & ResNet-101  & 43.7 \\ 
UperNet \cite{uper}  & ResNet-101  & 42.66 \\ 
DSSPN \cite{DSSPN}  & ResNet-101  & 43.68 \\ 
DM-Net \cite{dmsf}  & ResNet-101  & 45.5 \\
AC-Net \cite{ACnet}  & ResNet-101  & 45.9 \\
\bottomrule
\end{tabular}
\end{table}

\begin{table}
\centering
\caption{Instance Segmentation Models Performance on COCO test-dev 2017}
\label{instance_coco2017}
\begin{tabular}{llll}
\toprule
Method & Backbone & FPS  &  AP \\
\midrule
YOLACT-550 \cite{bolya2019yolact} & R-101-FPN & 33.5 & 29.8 \\
YOLACT-700 \cite{bolya2019yolact} & R-101-FPN & 23.8 & 31.2 \\
RetinaMask \cite{fu2019retinamask}  & R-101-FPN & 10.2 & 34.7 \\
TensorMask \cite{TensorMask}  & R-101-FPN & 2.6 & 37.1 \\ 
SharpMask \cite{SharpMask}  & R-101-FPN & 8.0 & 37.4 \\ 
Mask-RCNN \cite{mask_rcnn}  & R-101-FPN & 10.6 & 37.9 \\ 
CenterMask \cite{lee2020centermask}  & R-101-FPN & 13.2 & 38.3 \\
\bottomrule
\end{tabular}
\end{table}

\begin{table}
\centering
\caption{Panoptic Segmentation Models Performance on the MS-COCO
val dataset. $*$ denotes use of deformable convolution.}
\label{panoptic_coco2017}
\begin{tabular}{lll}
\toprule
Method & Backbone & PQ \\
\midrule
Panoptic FPN  \cite{pfpn} & ResNet-50  & 39.0 \\
Panoptic FPN  \cite{pfpn} & ResNet-101 & 40.3 \\
AU-Net \cite{seg_att4} & ResNet-50 &  39.6 \\
Panoptic-DeepLab \cite{cheng2019panoptic} & Xception-71 &  39.7 \\
OANet \cite{liu2019end} & ResNet-50 &  39.0 \\
OANet \cite{liu2019end} & ResNet-101 &  40.7 \\
AdaptIS \cite{sofiiuk2019adaptis} & ResNet-50 &  35.9 \\
AdaptIS \cite{sofiiuk2019adaptis} & ResNet-101 &  37.0 \\
UPSNet$^{*}$ \cite{xiong2019upsnet} & ResNet-50 & 42.5 \\
OCFusion$^{*}$ \cite{lazarow2020learning} & ResNet-50 & 41.3 \\
OCFusion$^{*}$ \cite{lazarow2020learning} & ResNet-101 & 43.0 \\
OCFusion$^{*}$ \cite{lazarow2020learning} & ResNeXt-101 & 45.7 \\
\bottomrule
\end{tabular}
\end{table}


\begin{table}
\centering
\caption{Performance of segmentation models on the NYUD-v2, and SUN-RGBD datasets, in terms of mIoU, and mean Accuracy (mAcc).}
\label{table_NYUv2}
\begin{tabular}{lrrrr}
\toprule
&  \multicolumn{2}{c}{NYUD-v2} & \multicolumn{2}{c}{SUN-RGBD}\\
\midrule
Method  & m-Acc  & m-IoU & m-Acc & m-IoU \\ \midrule
%
Mutex \cite{deng2015semantic}  & -   & 31.5 & - & - \\ 
MS-CNN \cite{rgbd_multiscale}  & 45.1 & 34.1  & - &  - \\  
FCN \cite{seg_fcn}  & 46.1  & 34.0 & - & - \\  
Joint-Seg \cite{mousavian2016joint}  & 52.3   & 39.2 & - &  - \\ 
SegNet \cite{segnet}  & -   & - & 44.76 & 31.84  \\ 
Structured Net \cite{seg_dsn2}  & 53.6   & 40.6 & 53.4 &  42.3 \\  
B-SegNet \cite{bayes_segnet}  & -   & - & 45.9 & 30.7  \\ 
3D-GNN \cite{3dgnn}  & 55.7  &  43.1 & 57.0 & 45.9 \\  
LSD-Net \cite{cheng2017locality}  & 60.7   & 45.9 & 58.0 & -  \\ 
RefineNet \cite{Refinenet}  &  58.9  & 46.5 & 58.5 & 45.9  \\ 
D-aware CNN \cite{wang2018depth}  & 61.1  &  48.4 & 53.5 & 42.0  \\  
RDFNet \cite{RDFNet}  & 62.8  &  50.1 & 60.1  &  47.7 \\  
G-Aware Net \cite{Geometry-Aware}  & 68.7  & 59.6 & 74.9 &  54.5  \\
MTI-Net \cite{Geometry-Aware}  & 68.7  & 59.6 & 74.9 &  54.5  \\
\bottomrule
\end{tabular}
\end{table}

To summarize the tabulated data, there has been significant progress
in the performance of deep segmentation models over the past 5--6
years, with a relative improvement of 25\%-42\% in mIoU on different datasets.
However, some publications suffer from lack of reproducibility for
multiple reasons---they report performance on non-standard
benchmarks/databases, or they report performance only on arbitrary
subsets of the test set from a popular benchmark, or they do not
adequately describe the experimental setup and sometimes evaluate the
model performance only on a subset of object classes. Most
importantly, many publications do not provide the source-code for
their model implementations. However, with the increasing popularity
of deep learning models, the trend has been positive and many research
groups are moving toward reproducible frameworks and open-sourcing
their implementations.

\section{Challenges and Opportunities}
\label{sec:challenges}

There is not doubt that image segmentation has benefited greatly from deep learning, but
several challenges lie ahead. We will next introduce some of the
promising research directions that we believe will help in further
advancing image segmentation algorithms.



\subsection{More Challenging Datasets}

Several large-scale image datasets have been created for semantic
segmentation and instance segmentation. However, there remains a need
for more challenging datasets, as well as datasets for different kinds
of images. For still images, datasets with a large number of objects
and overlapping objects would be very valuable. This can enable
training models that are better at handling dense object scenarios, as
well as large overlaps among objects as is common in real-world
scenarios.

With the rising popularity of 3D image segmentation, especially in
medical image analysis, there is also a strong need for large-scale 3D
images datasets. These datasets are more difficult to create than
their lower dimensional counterparts. Existing datasets for 3D image
segmentation available are typically not large enough, and some are
synthetic, and therefore larger and more challenging 3D image datasets can be very valuable.

\subsection{Interpretable Deep Models}

While DL-based models have achieved promising performance on
challenging benchmarks, there remain open questions about these
models. For example, what exactly are deep models learning? How should
we interpret the features learned by these models? What is a minimal
neural architecture that can achieve a certain segmentation accuracy
on a given dataset? Although some techniques are available to
visualize the learned convolutional kernels of these models, a
concrete study of the underlying behavior/dynamics of these models is
lacking. A better understanding of the theoretical aspects of these
models can enable the development of better models curated toward
various segmentation scenarios.

\subsection{Weakly-Supervised and Unsupervised Learning}

Weakly-supervised (a.k.a. few shot learning) \cite{zhou2018brief} and unsupervised learning \cite{jing2020self}
are becoming very active research areas. These techniques promise to
be specially valuable for image segmentation, as collecting labeled
samples for segmentation problem is problematic in many application
domains, particularly so in medical image analysis. The transfer
learning approach is to train a generic image segmentation model on a
large set of labeled samples (perhaps from a public benchmark), and
then fine-tune that model on a few samples from some specific target
application. Self-supervised learning is another promising direction
that is attracting much attraction in various fields. There are many
details in images that that can be captured to train a segmentation
models with far fewer training samples, with the help of
self-supervised learning. Models based on reinforcement learning could
also be another potential future direction, as they have scarcely
received attention for image segmentation. For example, MOREL
\cite{seg_RL1} introduced a deep reinforcement learning approach for
moving object segmentation in videos. 

\subsection{Real-time Models for Various Applications}

In many applications, accuracy is the most important factor; however,
there are applications in which it is also critical to have
segmentation models that can run in near real-time, or at least near
common camera frame rates (at least 25 frames per second). This is
useful for computer vision systems that are, for example, deployed in
autonomous vehicles. Most of the current models are far from this
frame-rate; e.g., FCN-8 takes roughly 100\,ms to process a
low-resolution image. Models based on dilated convolution help to
increase the speed of segmentation models to some extent, but there is
still plenty of room for improvement.

\subsection{Memory Efficient Models}

Many modern segmentation models require a significant amount of memory
even during the inference stage. So far, much effort has been directed
towards improving the accuracy of such models, but in order to fit
them into specific devices, such as mobile phones, the networks must
be simplified. This can be done either by using simpler models, or by
using model compression techniques, or even training a complex model
and then using knowledge distillation techniques to compress it into a
smaller, memory efficient network that mimics the complex model.

\subsection{3D Point-Cloud Segmentation}

Numerous works have focused on 2D image segmentation, but much fewer
have addressed 3D point-cloud segmentation. However, there has been an
increasing interest in point-cloud segmentation, which has a wide
range of applications, in 3D modeling, self-driving cars, robotics,
building modeling, etc. Dealing with 3D unordered and unstructured
data such as point clouds poses several challenges. For example, the
best way to apply CNNs and other classical deep learning architectures
on point clouds is unclear. Graph-based deep models can be a potential
area of exploration for point-cloud segmentation, enabling additional
industrial applications of these data.

\subsection{Application Scenarios}
In this section, we briefly investigate some application scenarios of recent DL-based segmentation methods, and some challenges ahead. Most notably, these methods have been successfully applied to segment satellite images in the field of remote sensing \cite{MA2019166}, including techniques for urban planning \cite{GAO2020295} or precision agriculture \cite{PAOLETTI2019279}. Remote sensing images collected by airborne platforms \cite{ABRAMS2019121} and drones \cite{KERKECH2020105446} have also been segmented using DL-based techniques, offering the opportunity to address important environmental problems such as those involving climate change. The main challenges of segmenting this kind of images are related to the very large dimensionality of the data (often collected by imaging spectrometers with hundreds or even thousands of spectral bands) and the limited ground-truth information to evaluate the accuracy of the results obtained by segmentation algorithms. 
Another very important application field for DL-based segmentation has been medical imaging \cite{TAJBAKHSH2020101693}. Here, an opportunity is to design standardized image databases that can be used to evaluate fast spreading new diseases and pandemics. Last but not least, we should also mention DL-based segmentation techniques in biology \cite{AMYAR2020104037} and evaluation of construction materials \cite{SONG2020106118}, which offer the opportunity to address highly relevant application domains but are subject to challenges related to the massive volume of the related image data and the limited reference information for validation purposes.

\section{Conclusions}
\label{sec:conclusions}

We have surveyed more than 100 recent image segmentation algorithms
based on deep learning models, which have achieved impressive
performance in various image segmentation tasks and benchmarks,
grouped into ten categories such as: CNN and FCN, RNN, R-CNN, dilated
CNN, attention-based models, generative and adversarial models, among
others. We summarized quantitative performance analyses of these
models on some popular benchmarks, such as the PASCAL VOC, MS COCO,
Cityscapes, and ADE20k datasets. Finally, we discussed some of the
open challenges and potential research directions for image
segmentation that could be pursued in the coming years.

\section*{Acknowledgments}
The authors would like to thank Tsung-Yi Lin from Google Brain, and Jingdong Wang and Yuhui Yuan from Microsoft Research Asia, for reviewing this work, and providing very helpful comments and suggestions.




\bibliographystyle{IEEEtran}




\begin{IEEEbiography}[{\includegraphics[width=0.95in,height=1.25in,clip,keepaspectratio]{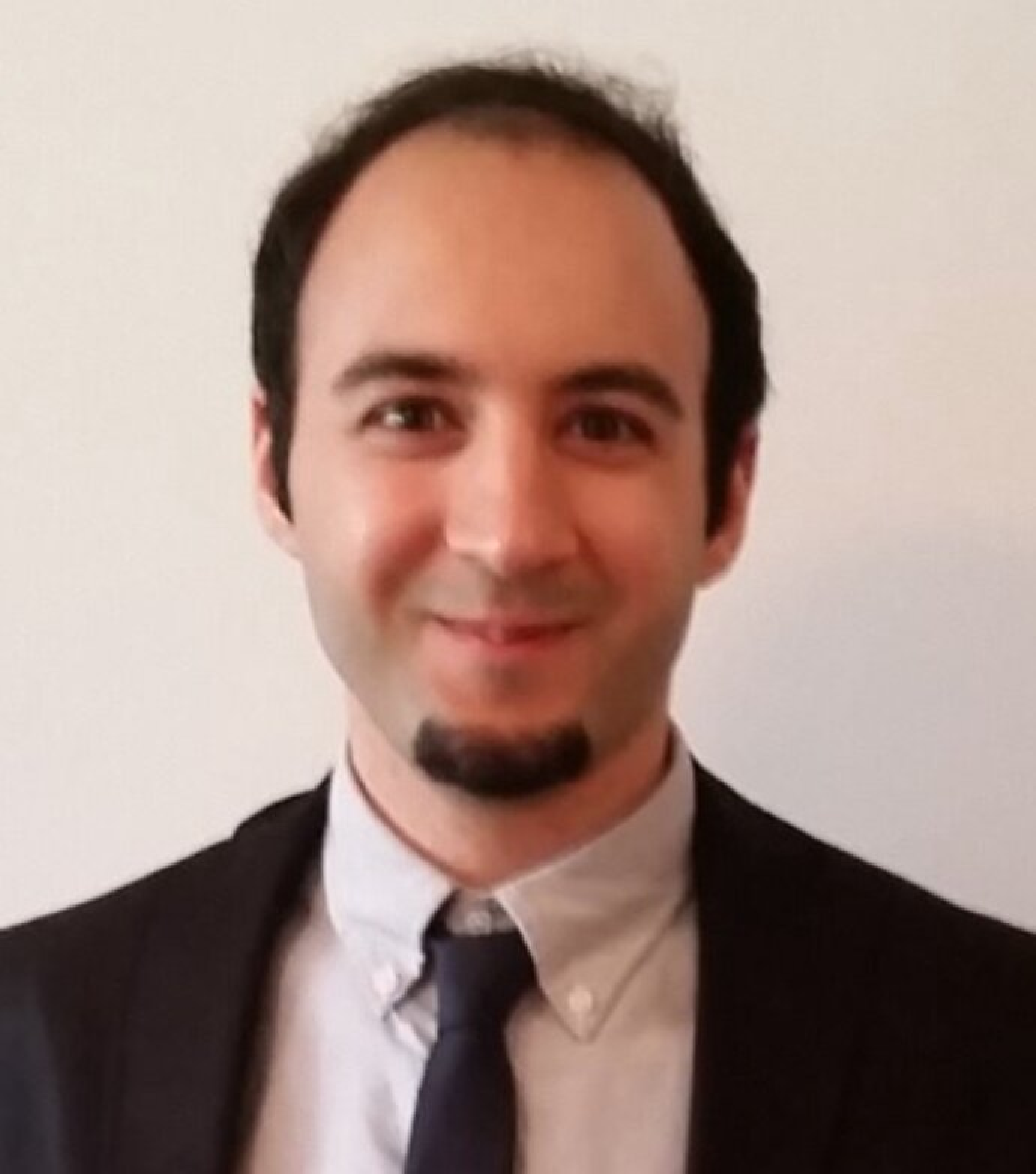}}]{Shervin Minaee} is a machine learning lead at Snapchat computer vision team. 
He received his PhD in Electrical Engineering and Computer Science from NYU, in 2018. 
His research interest includes computer vision, image segmentation, biometrics recognition, and applied deep learning.
He has published more than 40 papers and patents during his PhD.
He has  previously worked as a research scientist at Samsung Research, AT\&T Labs, Huawei Labs, and as a data scientist at Expedia group.
He is a reviewer for more than 20 computer vision related journals from IEEE, ACM, and Elsevier.
\end{IEEEbiography}

\vskip -5pt plus -1fil

\begin{IEEEbiography}[{\includegraphics[width=1in,height=1.25in,clip,keepaspectratio]{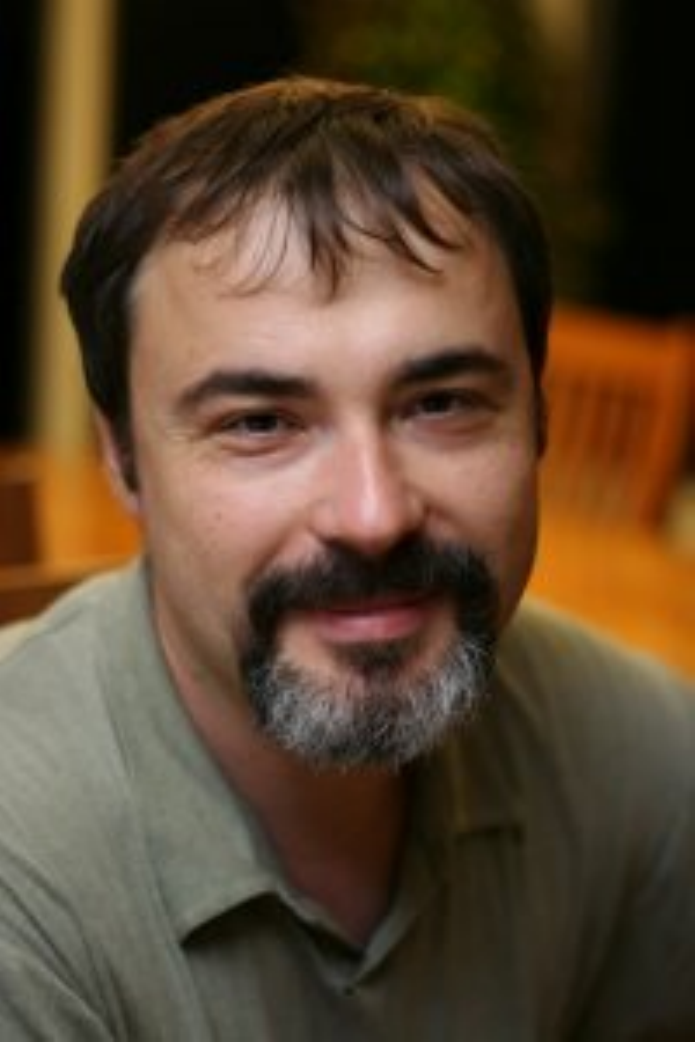}}]{Yuri Boykov} is a Professor at Cheriton School of Computer Science at the University of Waterloo. 
His research is concentrated in the area of computer vision and biomedical image analysis with focus on modeling and optimization for structured segmentation, restoration, registration, stereo, motion, model fitting, recognition, photo-video editing and other data analysis problems. He is an editor for the International Journal of Computer Vision (IJCV). His work was listed among 10 most influential papers in IEEE TPAMI (Top Picks for 30 years). In 2017 Google Scholar listed his work on segmentation as a "classic paper in computer vision and pattern recognition" (from 2006). In 2011 he received Helmholtz Prize from IEEE and Test of Time Award by the International Conference on Computer Vision. 
\end{IEEEbiography}

\vskip -1pt plus -1fil

\begin{IEEEbiography}[{\includegraphics[width=1in,height=1.25in,clip,keepaspectratio]{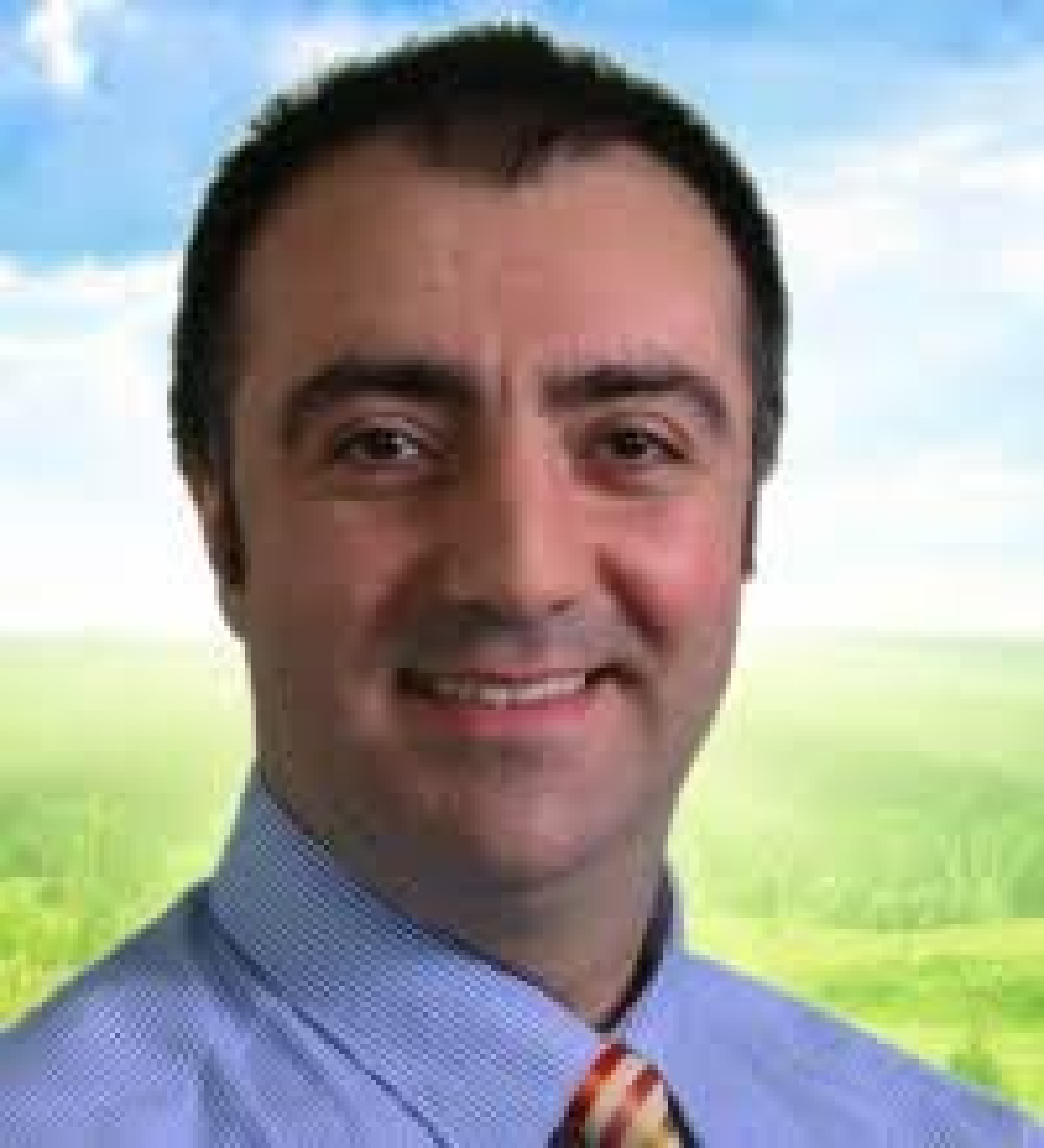}}]{Fatih Porikli} is an IEEE Fellow and a Senior Director at Qualcomm, San Diego. He was a full Professor in the Research School of Engineering, Australian National University and a Vice President at Huawei CBG Device; Hardware, San Diego until recently. He led the Computer Vision Research Group at NICTA, Australia and research projects as a Distinguished Research Scientist at Mitsubishi Electric Research Laboratories, Cambridge. He received his Ph.D. from New York University in 2002. He was the recipient of the R\&D 100 Scientist of the Year Award in 2006. He won six best paper awards, authored more than 250 papers, co-edited two books, and invented over 100 patents. 
He served as the General Chair and Technical Program Chair of many IEEE conferences and an Associate Editor of premier IEEE and Springer journals for the past 15 years.
\end{IEEEbiography}

\vskip -1pt plus -1fil

\begin{IEEEbiography}[{\includegraphics[width=1in,height=1.25in,clip,keepaspectratio]{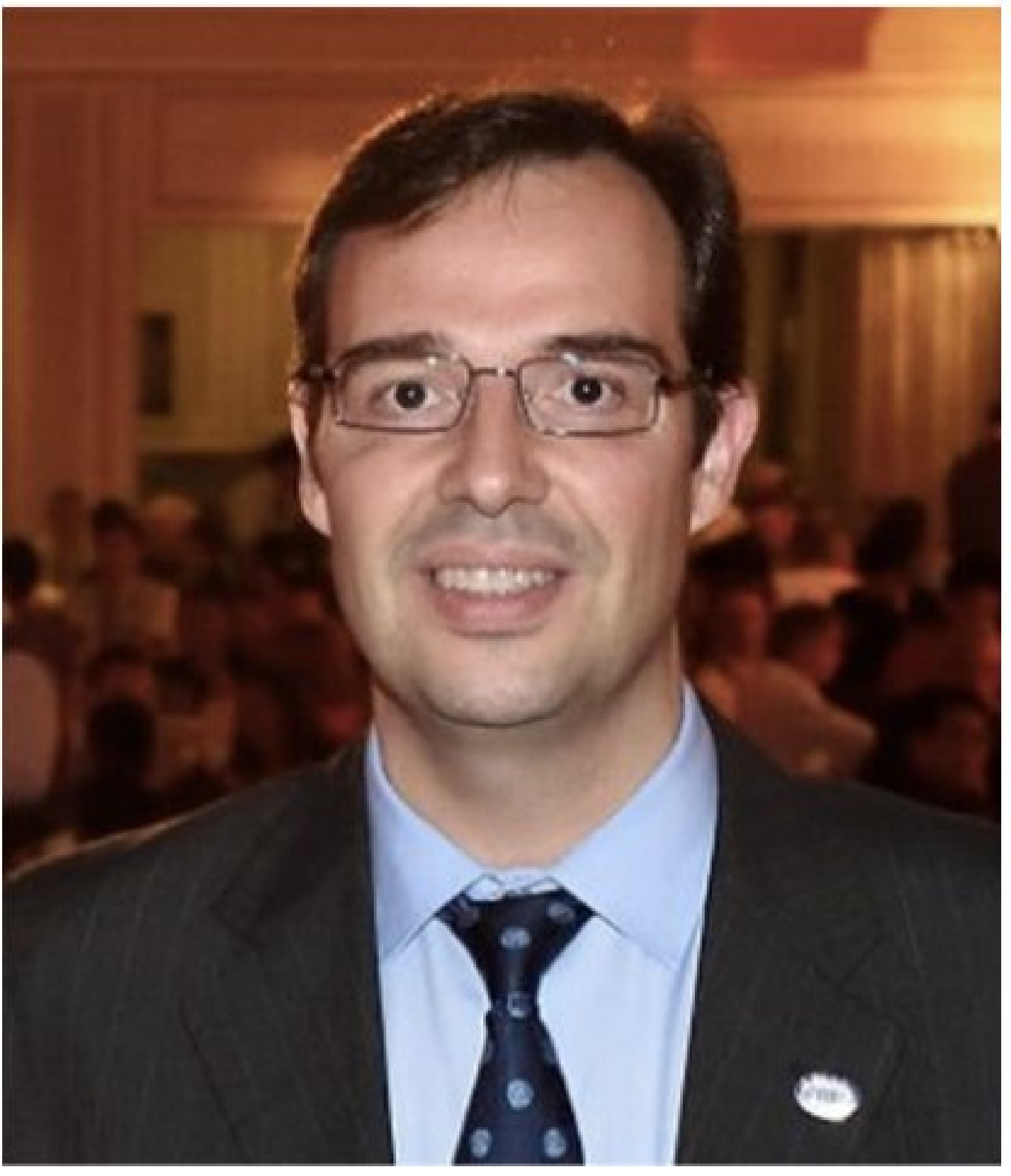}}]{Prof. Antonio Plaza} is  an IEEE Fellow  and a professor at the Department of Technology of Computers and Communications, University
of Extremadura, where he received the M.Sc. degree in 1999 and the PhD degree in 2002, both in Computer Engineering. 
He has authored more than 600 publications, including 300 JCR journal papers (more than 170 in IEEE journals), 24 book
chapters, and over 300 peer-reviewed conference proceeding papers. 
He is a recipient of the Best
Column Award of the IEEE Signal Processing Magazine in 2015, the 2013 Best Paper Award
of the JSTARS journal, and the most highly cited paper (2005-2010) in the Journal of
Parallel and Distributed Computing. 
He is included in the 2018 and 2019 Highly Cited Researchers List.
\end{IEEEbiography}

\vskip -5pt plus -1fil

\begin{IEEEbiography}[{\includegraphics[width=1in,height=1.25in,clip,keepaspectratio]{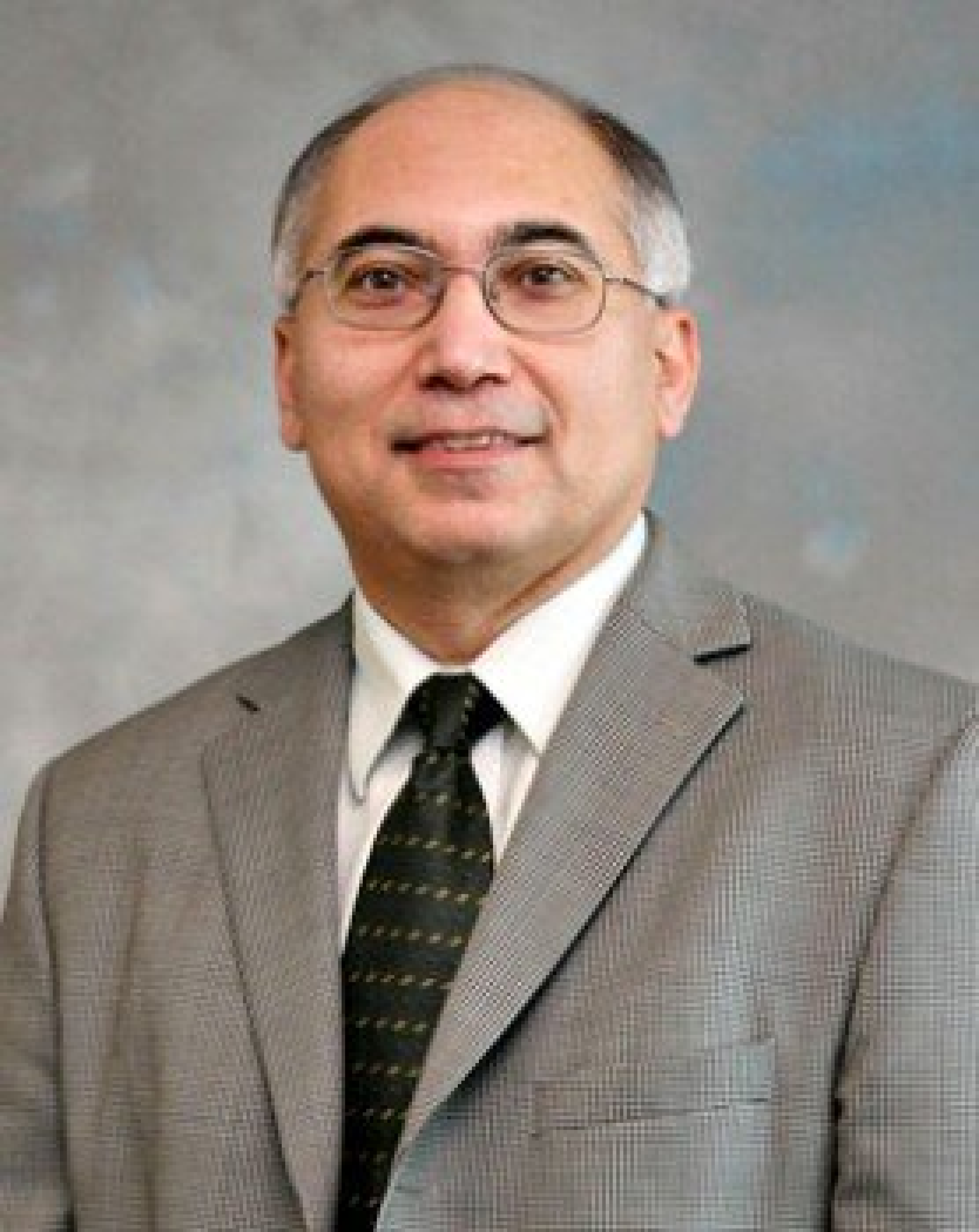}}]{Nasser Kehtarnavaz}
is a Distinguished Professor at the Department of Electrical and Computer Engineering at the University of Texas at Dallas, Richardson, TX. His research interests include signal and image processing, machine learning, and real-time implementation on embedded processors. He has authored or co-authored ten books and more than 390 journal papers, conference papers, patents, manuals, and editorials in these areas. He is a Fellow of SPIE, a licensed Professional Engineer, and Editor-in-Chief of Journal of Real-Time Image Processing.
\end{IEEEbiography}

\vskip -5pt plus -1fil

\begin{IEEEbiography}[{\includegraphics[width=1in,height=1.25in,clip,keepaspectratio]{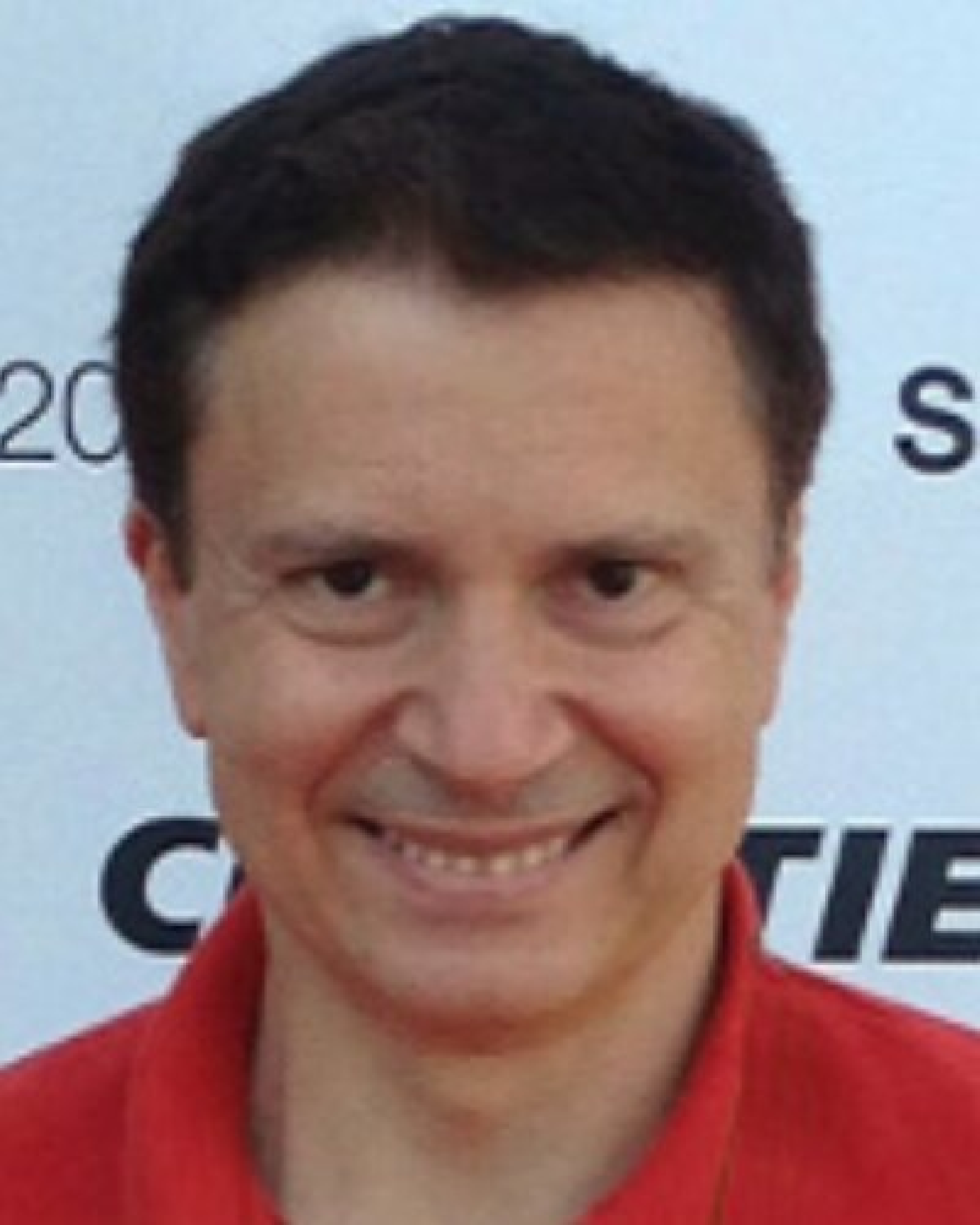}}]{Demetri Terzopoulos}
is a Distinguished Professor of Computer Science at the University of California, Los Angeles, where he directs the UCLA Computer Graphics \& Vision Laboratory.
He is also Co-Founder and Chief Scientist of \mbox{VoxelCloud}, Inc.
He received his PhD degree from Massachusetts Institute of Technology (MIT) in 1984. 
He is or was a Guggenheim Fellow, a Fellow of the ACM, IEEE, Royal Society of Canada, and Royal Society of London, and a Member of the European Academy of
Sciences, the New York Academy of Sciences, and Sigma Xi.
Among his many awards are an Academy Award from the Academy of Motion Picture Arts and Sciences for his pioneering work on physics-based computer animation, and the inaugural Computer Vision Distinguished Researcher Award from the IEEE for his pioneering and
sustained research on deformable models and their applications. ISI and other indexes have listed him among the most highly-cited authors
in engineering and computer science, with more than 400 published research papers and several volumes.
Previously, he was Professor of Computer Science and Professor of Electrical and Computer Engineering at the University of Toronto. 
Before becoming an academic in 1989, he was a Program Leader at Schlumberger corporate research centers in California and Texas.
\end{IEEEbiography}

\end{document}